\documentclass[12pt]{article}

\usepackage{scicite}
\usepackage{times}

\usepackage{subcaption} 

\usepackage{graphicx}

\usepackage[utf8]{inputenc} 
\usepackage[T1]{fontenc}    
\usepackage{hyperref}       
\usepackage{url}            
\usepackage{booktabs}       
\usepackage{amsfonts}       
\usepackage{nicefrac}       
\usepackage{microtype}      
\usepackage{diagbox}
\usepackage[dvipsnames,table]{xcolor}
\usepackage{float}
\usepackage{graphicx}
\usepackage{titlesec}
\usepackage{enumitem}
\usepackage{xspace}
\usepackage{tabularx}
\usepackage{algorithm}
\usepackage{algorithmicx}
\usepackage{algpseudocode}
\usepackage{amsmath}
\usepackage[normalem]{ulem}
\usepackage{multirow}
\usepackage{tcolorbox}
\usepackage{booktabs,tikz}
\usepackage[font=small]{caption}

\newcommand{\reled}{R-NaD\xspace}
\newcommand{\agent}{\emph{DeepNash}\xspace}

\newcommand\tikzmark[1]{\tikz[remember picture] \node (#1) {};}

\newcolumntype{M}[1]{>{\centering\arraybackslash}m{#1}}

\newenvironment{sciabstract}{%
\begin{quote} \bf}
{\end{quote}}

\title{Mastering the Game of Stratego with Model-Free Multiagent Reinforcement Learning} 

\date{}
\usepackage{authblk}
\author[,1,$\ddag$]{Julien~Perolat\footnote{corresponding authors : perolat@deepmind.com, bartdv@deepmind.com and karltuyls@deepmind.com}}
\author[$*$,1,$\ddag$]{Bart~de~Vylder}

\author[1]{Daniel~Hennes}
\author[1]{Eugene~Tarassov}
\author[1]{Florian~Strub}
\author[1]{Vincent~de~Boer\footnote{Independent consultant to DeepMind}}
\author[1]{Paul~Muller}
\author[1]{Jerome~T.~Connor}
\author[1]{Neil~Burch}
\author[1]{Thomas~Anthony}
\author[1]{Stephen~McAleer}
\author[1]{Romuald~Elie}
\author[1]{Sarah~H.~Cen}
\author[1]{Zhe~Wang}
\author[1]{Audrunas~Gruslys}
\author[1]{Aleksandra~Malysheva}
\author[1]{Mina~Khan}
\author[1]{Sherjil~Ozair}
\author[1]{Finbarr~Timbers}
\author[1]{Toby~Pohlen}
\author[1]{Tom~Eccles}
\author[1]{Mark~Rowland}
\author[1]{Marc~Lanctot}
\author[1]{Jean-Baptiste~Lespiau}
\author[1]{Bilal~Piot}
\author[1]{Shayegan~Omidshafiei}

\author[1]{Edward~Lockhart}
\author[1]{Laurent~Sifre}
\author[1]{Nathalie~Beauguerlange}
\author[1]{Remi~Munos}
\author[1]{David Silver}
\author[1]{Satinder~Singh}
\author[1]{Demis~Hassabis}
\author[$*$,1,$\ddag$]{Karl~Tuyls}
\affil[1]{DeepMind}
\affil[$\ddag$]{\normalsize{Shared lead authors}}

\begin{document} 


\baselineskip24pt


\maketitle 



\begin{sciabstract}
We introduce \agent, an autonomous agent capable of learning to play the imperfect information game Stratego\footnote{Stratego is a trademark of Jumbo Diset Group, and is used in this publication for information purposes only.} from scratch, up to a human expert level. Stratego is one of the few iconic board games that Artificial Intelligence (AI) has not yet mastered. This popular game has an enormous game tree on the order of $10^{535}$ nodes, i.e., $10^{175}$ times larger than that of Go. It has the additional complexity of requiring decision-making under imperfect information, similar to Texas hold’em poker, which has a significantly smaller game tree (on the order of $10^{164}$ nodes). Decisions in Stratego are made over a large number of discrete actions with no obvious link between action and outcome.
Episodes are long, with often hundreds of moves before a player wins, and situations in Stratego can not easily be broken down into manageably-sized sub-problems as in poker.
For these reasons, Stratego has been a grand challenge for the field of AI for decades, and existing AI methods barely reach an amateur level of play. \agent uses a game-theoretic, model-free deep reinforcement learning method, without search, that learns to master Stratego via self-play. The Regularised Nash Dynamics (R-NaD) algorithm, a key component of \agent, converges to an approximate Nash equilibrium, instead of ‘cycling’ around it, by directly modifying the underlying multi-agent learning dynamics.
\agent beats existing state-of-the-art AI methods in Stratego and achieved a yearly (2022) and all-time top-3 rank on the Gravon games platform, competing with human expert players.
\end{sciabstract}

\section{Introduction}
\begin{figure}[t!]
    \begin{subfigure}[b]{\textwidth}
    \begin{subfigure}[t][][b]{.33\textwidth}
        {
        \centering
        \includegraphics[width=\linewidth]{figures_final/boards/deploy_final}
        \caption*{\textbf{Phase 1:} Private deployment}
        }
    \end{subfigure}
    \hfill
    \begin{subfigure}[t][][b]{.33\textwidth}
        {
        \centering
        \includegraphics[width=\linewidth]{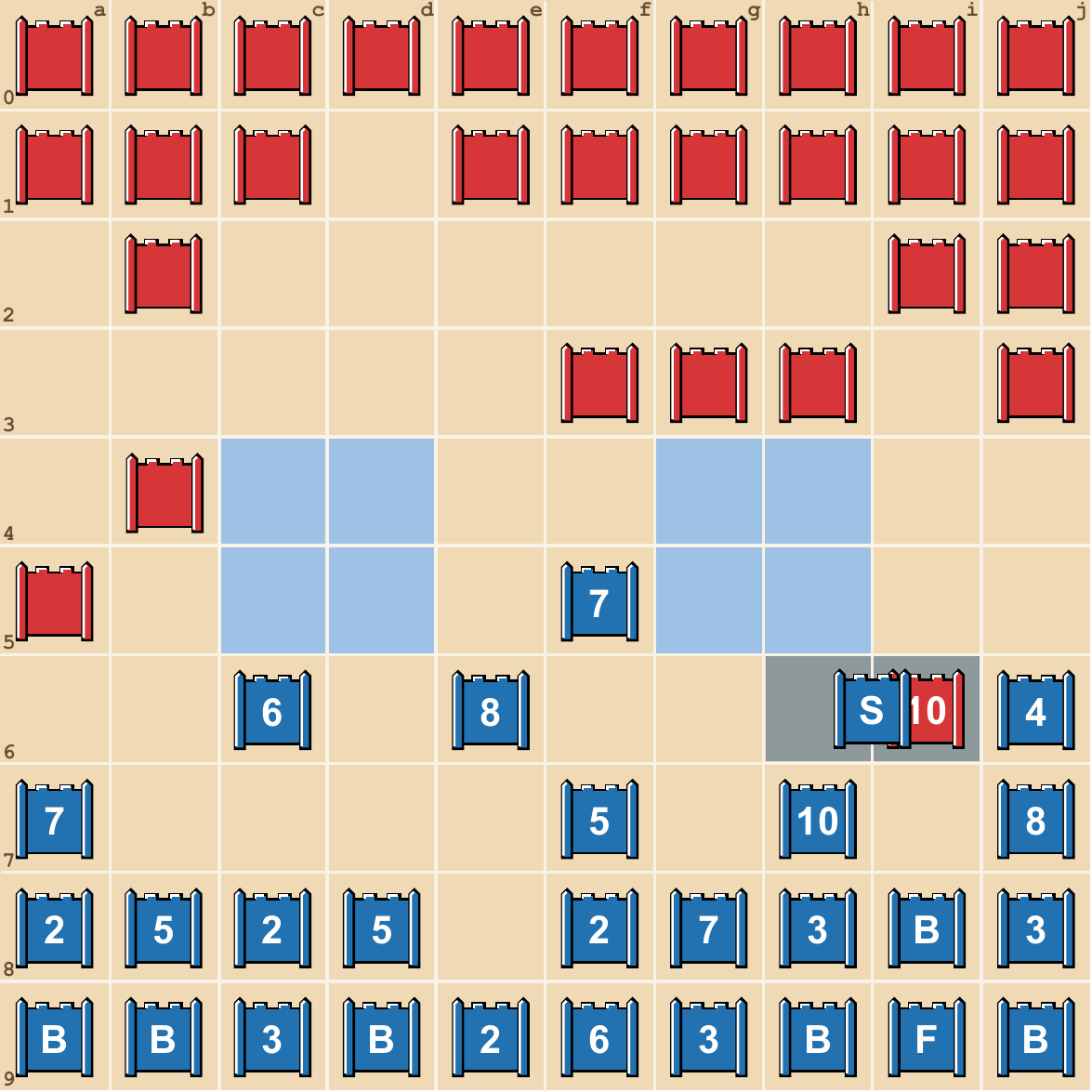}
        \caption*{\textbf{Phase 2:} Game play}
        }
    \end{subfigure}
     \hfill
    \begin{subtable}[t][][b]{0.3\textwidth}
    \vskip 0.15cm
            \centering
            \scriptsize
            \setlength\tabcolsep{4pt}
            \begin{tabular}{p{0.2cm}M{0.4cm}l}
            \tikzmark{a} &   \includegraphics[width=0.3cm]{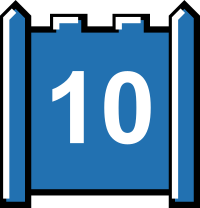} & Marshal  \tikzmark{d}\\
                         & \includegraphics[width=0.3cm]{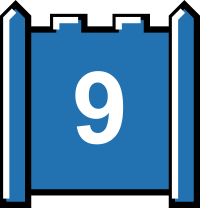}    & General             \\
                         & \includegraphics[width=0.3cm]{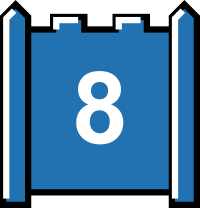}    & Colonel             \\
                         & \includegraphics[width=0.3cm]{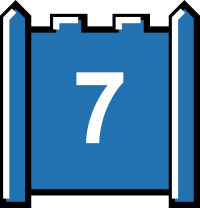}    & Major               \\
                         & \includegraphics[width=0.3cm]{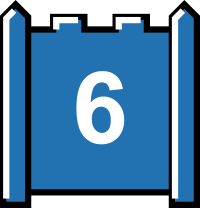}    & Captain             \\
                         & \includegraphics[width=0.3cm]{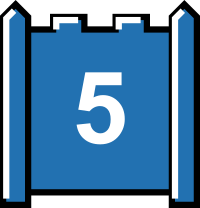}    & Lieutenant          \\
                         & \includegraphics[width=0.3cm]{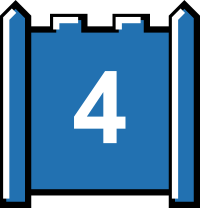}    & Sergeant             \\
                         & \includegraphics[width=0.3cm]{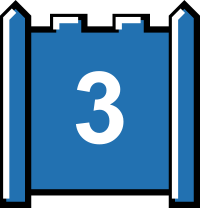}    & Miner: diffuses Bombs  \\
                         &  \includegraphics[width=0.3cm]{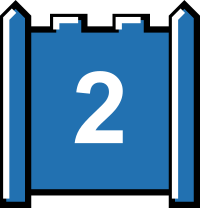}   & Scout: long range move \\
             \tikzmark{b}  & \includegraphics[width=0.3cm]{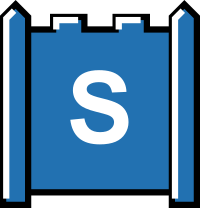}  & Spy: defeats Marshal \; \tikzmark{c}\\
                         \arrayrulecolor{black!30}\midrule
                         & \includegraphics[width=0.3cm]{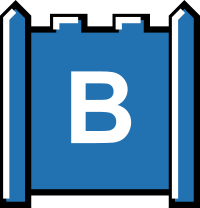}   & Bomb: immobile; only captured by  Miner\\
                         &\includegraphics[width=0.3cm]{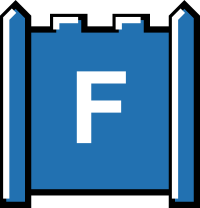}   & Flag: immobile, game over when captured
            \end{tabular}
            \caption*{Piece types}
            \tikz[remember picture,overlay] \draw[<-,  >=latex, red!20!white, line width=3pt] (a.center -| b.center) to node[black,rotate=270,fill=white]{Increasing strength} (b.center);
            \tikz[remember picture,overlay] \draw[<->,  >=latex] (c.center) [bend right=70]to node[black,sloped,fill=white]{Captures} (d.center);
        \end{subtable}
        \caption{Stratego is a two-player board game where each player aims to capture the opponent's flag. To do so, they each have 40 pieces  of diverse strengths. The game starts with the deployment phase, where both players secretly position their pieces on the board. In a second game-play phase, the players take turns moving pieces. When two pieces are in the same location, they are revealed, and the weaker piece is removed, or both if they have the same strength. When the weakest movable piece, the Spy, attacks the 10, however, it wins and the 10 is captured. The players have only a partial view on the opponent's pieces: seeing their position but not their type. The complete rules~\cite{stratego_rules} are defined by the International Stratego Federation.}
        \label{fig:strategogamemechanics}
        \vspace{1.5em}
    \end{subfigure}
    %
    \begin{subfigure}[b]{\textwidth}
        \includegraphics[width=1\textwidth,trim={0 1.5cm 0 0},clip]{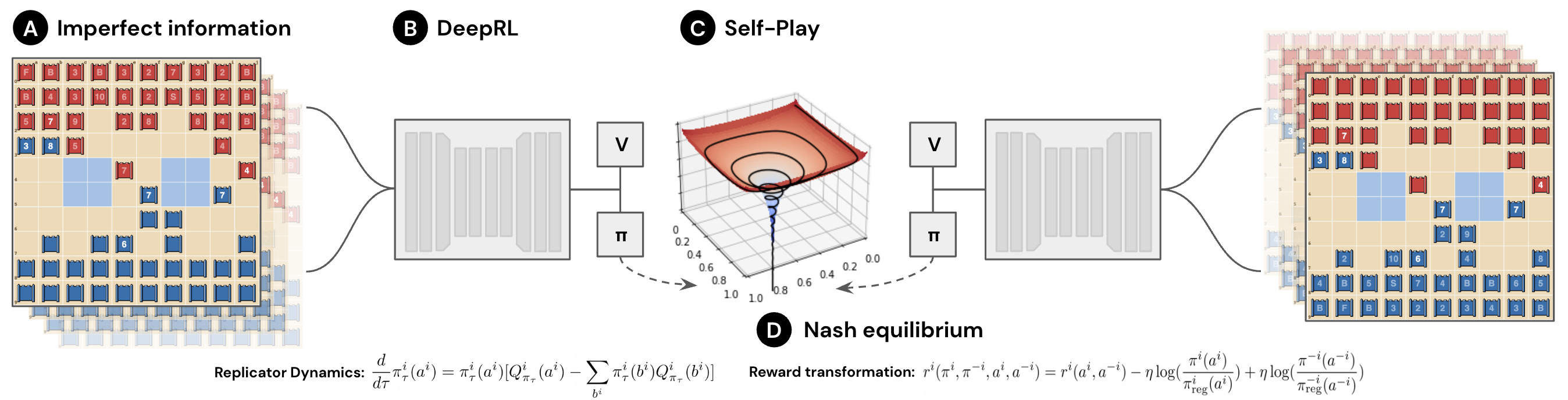}
        \caption*{{\scriptsize \centerline{\textbf{Replicator dynamics:}     $\frac{d}{d\tau}\pi^{i}_{\tau}(a^i) = \pi^{i}_{\tau}(a^i)\big[Q^{i}_{\pi_\tau}(a^i) - \sum_{b^i} \pi^{i}_{\tau}(b^i)Q^{i}_{\pi_\tau}(b^i) \big]$}     \centerline{\textbf{Reward transformation:}    $r^{i}(\pi^{i}, \pi^{-i}, a^{i}, a^{-i})=r^{i}(a^{i}, a^{-i}) - \eta \log\big(\frac{\pi^{i}(a^{i})}{\pi^{i}_{\text{reg}}(a^{i})}\big) + \eta \log\big(\frac{\pi^{-i}(a^{-i})}{\pi^{-i}_{\text{reg}}(a^{-i})}\big)$}}     }
        \caption{An overview of the DeepNash approach. DeepNash is an autonomous agent that learns to play the imperfect information game Stratego (A). It learns a policy represented by a deep neural network (B) through self-play from scratch (C) in order to converge to a Nash equilibrium (D).}
        \label{fig:overviewdeepnash}
    \end{subfigure}
    \caption{The Stratego game (a) and an overview of the \agent approach (b).}
\end{figure}

Progress in Artificial Intelligence (AI) has been measured via mastery of board games since the inception of the field~\cite{shannon:philosophicalmag1950}. Board games allow us to gauge and evaluate how humans and machines develop and execute strategies in a controlled environment. The ability to plan ahead has been at the heart of successes in AI for decades in perfect information games such as chess, checkers, shogi and Go, as well as in imperfect information games such as poker and Scotland Yard~\cite{Campbell:2002aa,SilHub18General,schrittwieser2020mastering,moravvcik2017deepstack,brown2018superhuman,brown2019superhuman,PlayerofGames21}.  
For many years the Stratego board game has constituted one of the next frontiers of AI research. For a visualization of the game phases and game mechanics see Figure \ref{fig:strategogamemechanics}. The game poses two key challenges. First,
the game tree of Stratego has $10^{535}$ possible states, which is larger than both no-limit Texas hold'em poker, a well-researched imperfect information game with $10^{164}$ states~\cite{stratego2010afc}, and the game of Go, which has $10^{360}$ states~\cite{stratego2010afc}.
Second, acting in a given situation in Stratego requires reasoning over $10^{66}$ possible deployments at the start of the game for each player, whereas poker has only $10^3$ possible pairs of cards~\cite{johanson2013measuring}. Perfect information games like Go and chess do not have a private deployment phase, therefore avoiding the complexity this challenge poses in Stratego.
Currently it is not possible to use state-of-the-art model-based perfect information planning techniques, nor state-of-the-art imperfect information search techniques that break down the game into independent situations \cite{brown2019superhuman,PlayerofGames21}.

For these reasons, Stratego provides a challenging benchmark for studying strategic interactions at an unparalleled scale. As in most board games, Stratego tests our ability to make relatively slow, deliberative, and logical decisions sequentially. Most recent successes in large imperfect information games have been achieved in real-time strategy games such as Starcraft, Dota and Capture the flag~\cite{vinyals2019grandmaster, jaderberg2019human, berner2019dota} in which most decisions must be made quickly and instinctively, and are of a continuous-time nature. Stratego is a game in which little progress has been achieved by the AI research community due to many complex aspects of the game's structure. Successes in the game have been limited, with artificial agents only able to play at a level comparable to a human amateur, see e.g. \cite{TreijtelR01,deboer2007,BoerRW08,Schadd09,Satz11,Redeca18,mcaleer2020pipeline}. Developing intelligent agents that learn end-to-end to make optimal decisions under imperfect information in Stratego, from scratch, without human demonstration data, remained one of the grand challenges of AI research.

In this work we introduce \agent, an agent that learns to play Stratego in self-play in a model-free manner without human demonstration, beating previous state-of-the-art AI agents and achieving expert human-level performance in the most complex variant of the game, Stratego Classic. At the core of \agent is a principled, model-free reinforcement learning algorithm called \emph{Regularized Nash Dynamics} (\reled). 
\agent combines \reled with a deep neural network architecture and converges to an $\epsilon$-Nash equilibrium, which means it learns to play at a highly competitive level, and is robust against opponents that try to exploit it. All games of imperfect information possess a Nash equilibrium in mixed strategy~\cite{Nash51}, assigning a mixed (or stochastic) strategy for all the players in which no player benefits from deviating from their strategy as long as no other player deviates.
While it is sufficient to take deterministic decisions that maximize the value of the equilibrium strategy in turn-taking two-player zero-sum games of full information, this approach is theoretically unsound when dealing with imperfect information games.
In such games, other tactics need to be deployed, which better reflect decision-making processes in the real world. As von Neumann described it "real life consists of bluffing, of little tactics of deception, of asking yourself what is the other man going to think I mean to do."\cite{vonneumann1947}. Figure \ref{fig:overviewdeepnash} illustrates a high-level overview of the \agent approach.

We lay out our new model-free reinforcement learning method \agent, and systematically evaluate its performance against various state-of-the-art Stratego bots and human expert players on the Gravon games platform~\cite{gravon_site}. \agent convincingly beats all current state-of-the-art bots that have been developed to play Stratego with a win rate of over $97\%$ and achieves a highly competitive level of play with human expert Stratego players on Gravon, where it ranks among the top $3$ players, both on the annual (2022) and all-times leaderboards, with a win rate of $84\%$. As such, it is the first time an AI algorithm is able to learn to play at a human-expert level in a complex board game without deploying any search method in the learning algorithm, and the first time an AI achieves human-expert level in the game of Stratego.

\section{Methods}
\agent takes an end-to-end learning approach to solve Stratego, by incorporating the learning of the deployment phase, i.e., putting the pieces tactically on the board at the start of a game (see Figure \ref{fig:strategogamemechanics}), in the learning of the game-play phase, using an integrated deep RL and game-theoretic approach.
The agent's purpose is to learn an approximate Nash equilibrium through self-play. A Nash equilibrium guarantees that the agent will perform well, even against a worst case opponent. Designing a strategy to be robust in the worst case is typically a good choice to play well against humans in two-player zero-sum games (see e.g. \cite{MoravcikSBLMBDW17,brown2018superhuman,brown2019superhuman}), as a Nash equilibrium guarantees an unexploitable agent, and thus the best possible worst-case performance. In perfect information games, search techniques aided by reinforcement learning, i.e. model-based learning techniques, have provided state-of-the-art superhuman bots in Go and chess \cite{SilverHMGSDSAPL16,SilHub18General}. However, searching for a Nash equilibrium in imperfect information games requires estimating private information of the opponent from public states~\cite{moravvcik2017deepstack,brown2020combining,PlayerofGames21}. Given the vast number of such possible private configurations in a public state, Stratego computationally challenges all existing search techniques as the search space becomes intractable. We therefore chose an orthogonal route in this work, without search, and propose a new method that combines model-free reinforcement learning in self-play with a game-theoretic algorithmic idea, \emph{Regularized Nash Dynamics} (\reled). The model-free part implies that we don't build an explicit opponent model tracking belief space (calculating a likelihood of the opponent's state), and the game-theoretic part is based on the idea that by modifying the dynamical system underpinning our reinforcement-learning approach we can steer the learning behavior of the agent in the direction of the Nash equilibrium. The main advantage of this combined approach is that we do not need to explicitly model private states from public ones. A complex challenge, on the other hand, is to scale up this model-free reinforcement learning approach with \reled to make self-play competitive against human expert players in Stratego, which has not been achieved to date. This combined \agent approach is illustrated in Figure \ref{fig:overviewdeepnash}.

In the following subsections, we will use elementary concepts from game theory, and refer the unfamiliar reader for more details to the background section in the supplementary material.

\subsection{Learning approach}
We learn a Nash equilibrium in Stratego through self-play and model-free reinforcement learning. The idea of combining model-free RL and self-play has been tried before, but it has been empirically challenging to stabilize such learning algorithms when scaling up to complex games, as for example Capture the flag, Dota and StarCraft~\cite{alphastar,jaderberg2019human,OpenAI_dota}. Some empirical work manages to stabilize the learning either by training against past versions of the agent~\cite{alphastar,jaderberg2019human,OpenAI_dota}, or by adding reward-shaping~\cite{OpenAI_dota,jaderberg2019human} or expert data~\cite{alphastar} in the training algorithm. While these are helpful tricks, such approaches lack theoretical foundations, remain hard to tune and do not easily generalize to new games. Furthermore, in a game like Stratego, it is difficult to define a loss whose minimization would converge to a Nash equilibrium without introducing prohibitive computational obstacles at large scale. For instance, minimizing the exploitability~\cite{ijcai2019-0066}, a well-known quantity that measures the distance to a Nash equilibrium, requires estimating an agent's best response during training, which is computationally intractable in Stratego.
However, it is possible to define a learning update rule that induces a dynamical system for which there exists a so-called Lyapunov function. This function can be shown to decrease during learning and as such guarantees convergence to a fixed point. This is the central idea behind the \reled algorithm, and the successful recipe for \agent, which scales this approach using a deep neural network.

\subsection{Regularized Nash Dynamics algorithm}

\begin{figure}[t]
    \begin{minipage}[b]{.5\textwidth}
        \begin{subtable}[b]{\textwidth}
            \centering
            \footnotesize
            \begin{tabular}{cr|cc}
                &\multicolumn{1}{c}{} & \multicolumn{2}{c}{\emph{Player 2}}\rule[-2ex]{0pt}{0pt}    \\
                &   & Head: $H$    & Tail: $T$ \\
                \cline{2-4}
                \multirow{2}{*}{\emph{Player 1}}& Head: $H$ & $1$   & $-1$ \\
                 & Tail: $T$  & $-1$  &  $1$ \\
            \end{tabular}
            \caption{Matching pennies}
            \label{fig:reled_illustration:pennies}
        \end{subtable}
        \vskip 1cm
      \begin{subfigure}[b]{\textwidth}
            \footnotesize
            \begin{tcolorbox}
                \centerline{\textbf{\reled Iteration}}
                \vspace{0.5em}
                Start with an arbitrary regularization policy: $\pi_{0,\text{reg}}$
                \begin{enumerate}[leftmargin=*]
                    \item \underline{Reward transformation:} Construct the transformed game with: $\pi_{n,\textrm{reg}}$
                    \item \underline{Dynamics:} Run the replicator dynamics until convergence to: $\pi_{n,\textrm{fix}}$
                    \item \underline{Update:} Set the regularization policy: \\
                    \centerline{$\pi_{n+1,\textrm{reg}} = \pi_{n,\textrm{fix}}$}
                \end{enumerate}
                Repeat steps until convergence
            \end{tcolorbox}
        \caption{Algorithmic steps}
        \label{fig:reled_illustration:algo}
        \end{subfigure}
    \end{minipage}
    \begin{minipage}[b]{0.5\textwidth}
     \begin{subfigure}[b]{\textwidth}
        \centering
        \includegraphics[width=\linewidth]{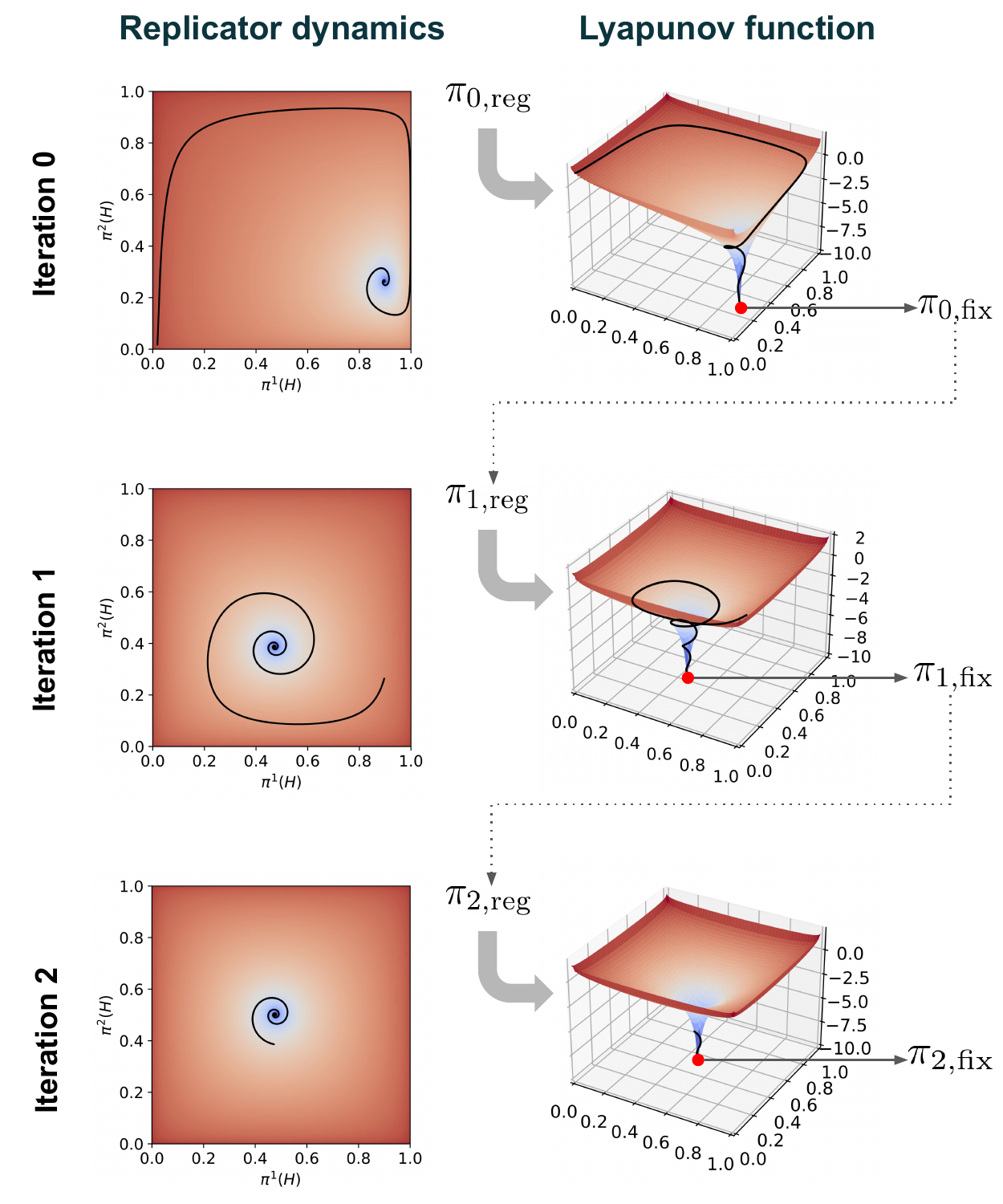}
        \caption{Dynamics and Lyapunov function}
        \label{fig:reled_illustration:traj}
    \end{subfigure}
    \end{minipage}
    \caption{The \reled learning algorithm illustrated with the  matching pennies game}

\end{figure}

The \reled learning algorithm used in \agent is based on the idea of regularization~\cite{Tuyls03,Tuyls06,TuylsHNM03,perolat2020poincar,piliouras2017learning,mertikopoulos2017cycles,McMahan11a} for convergence purposes, which we briefly first explain in the context of zero-sum two-player normal form games (illustrated on the matching pennies game).\footnote{An NFG is an abstraction of a decision-making situation involving more than one agent. Each agent needs to simultaneously take an action, after which they receive a game reward, and the game starts a new iteration of the same situation.}
\reled relies on three key steps (see also Figure \ref{fig:reled_illustration:algo}):

First a \emph{reward transformation} step is performed based on a regularization policy $\pi_{\textrm{reg}}$ which induces a modified game with rewards: $r^i(\pi^i, \pi^{-i},a^i, a^{-i}) = r^i(a^i, a^{-i}) - \eta \log(\frac{\pi^i(a^i)}{\pi^i_{\textrm{reg}}(a^i)}) + \eta \log(\frac{\pi^{-i}(a^{-i})}{\pi^{-i}_{\textrm{reg}}(a^{-i})})$, with $\eta > 0$ a regularization parameter and $i$ the player index ($i\in [1, 2]$). Note that this transformed reward is policy-dependent. 

Second, in the \emph{dynamics} step we let the system evolve according to the  replicator dynamics system~\cite{Zeeman80,Zeeman81,smith,BloembergenTHK15} on this modified game. Replicator dynamics are a descriptive learning process from evolutionary game theory, equivalent to RL algorithms~\cite{Tuyls03,BloembergenTHK15}, that are also known as Follow the Regularized Leader~\cite{McMahan11a}, and defined as follows: 
$$  \frac{d}{d\tau}\pi^i_\tau(a^i) = \pi^i_\tau(a^i)[Q^i_{\pi_\tau}(a^i)-\sum\limits_{b^i}\pi^i_\tau(b^i)Q^i_{\pi_\tau}(b^i)]$$
with $Q^i_{\pi_\tau}(a^i)$ the quality or fitness of an action. These dynamics reinforce the probability of taking actions with high fitness (relative to other actions). Thanks to the reward transformation this system has a unique fixed point $\pi_\textrm{fix}$ and convergence to it is guaranteed, which can be proven by the Lyapunov function $H_{\pi_{\textrm{fix}}}(\pi) = \sum \limits_{i=1}^{2} \sum \limits_{a^i \in A^i} \pi^i_{\textrm{fix}}(a^i)\log\left(\frac{\pi^i_{\textrm{fix}}(a^i)}{\pi^i(a^i)}\right)$~\cite{perolat2020poincar}. however, this fixed point is not yet a Nash equilibrium of the original game. 

In the final \emph{update} step, the fixed point obtained is used as the regularization policy for the next iteration. These three steps are applied repeatedly, generating a sequence of fixed points which can be proven to converge to a Nash equilibrium of the original (unmodified) game~\cite{perolat2020poincar}. 
Figure ~\ref{fig:reled_illustration:traj} illustrates the \reled algorithm on the two-player matching pennies game (with the payoff table in Figure \ref{fig:reled_illustration:pennies}).
The first iteration starts from  $\pi^i_{0,\textrm{reg}}[H, T] = [0.999, 0.001]$, ($\eta=0.2$) and the replicator dynamics converge to $\pi^0_{0,\textrm{fix}}[H, T] = [0.896, 0.104]$ and $\pi^1_{0,\textrm{fix}}[H, T] = [0.263, 0.737]$. The right figure shows the evolution of the logarithm of the Lyapunov function and illustrates it decreases while learning. Three iterations of \reled are shown.

\subsection{DeepNash: \reled at scale}

\agent consists of three components: (1) a core training component \reled, the model-free RL algorithm presented above, implemented using a deep convolutional network, (2) fine-tuning of the learnt policy to reduce the residual probabilities of taking highly improbable actions and, (3) test-time post-processing to filter out low probability actions and clear mistakes.  

We start by concisely laying out some essential background information on imperfect information games necessary to understand how \reled is scaled to a deep learning model. Then we continue to unpack the three algorithmic steps of \reled and summarize how they are implemented in the neural architecture. For a detailed description we refer to the supplemental material.

\subsubsection{Imperfect information games}

In a two-player zero-sum imperfect information game, two players (player $i=1$ or $i=2$) sequentially interact in turns. At turn $t$ the players receive a reward signal $(r^1_t, r^2_t)$ and the current player $i=\psi_t$ observes the game state through an observation $o_t$ and selects an action $a_t$ according to a parameterized policy function $\pi(.|o_t)$. In model-free reinforcement learning the trajectories $\mathrm{T} = [(o_t, a_t, (r^1_t, r^2_t),\pi(.|o_t)), \psi_t]_{0\leq t <t_{\textrm{max}}}$ are the only data the agent will leverage to learn the parameterized policy.

\subsubsection{Model-free Reinforcement Learning with Regularized Nash Dynamics}
\agent scales the \reled algorithm by using deep learning architectures. It carries out the same three algorithmic steps as before in NFGs: (1) the \emph{reward transformation step}, which modifies the reward, (2) the \emph{dynamics step} which allows for convergence to a fixed point, and (3) the \emph{update step} in which the algorithm updates the policy that defines the regularization function.

\paragraph{Neural architecture and observation representation:}
\agent's network consists of the following components: a U-Net torso with residual blocks and skip-connections~\cite{ronneberger2015u,he2016deep}, and four heads which are smaller replicas of the torso augmented with final layers to generate an output of the appropriate shape. The first \agent head outputs the value function as a scalar, while the three remaining heads encode the agent's policy by outputting a probability distribution over its actions at deployment and during gameplay. The agent  architecture is described in detail in the supplementary material.

The observation is encoded as a spatial tensor consisting of the following components: \agent's own pieces, the publicly available information about both the opponent's and 
\agent's pieces and an encoding of the $40$ last moves. This public information  represents the types each piece can still have given the history of the game. In total, the observation contains $82$ stacked frames encoded in a single tensor. The structure of this observation tensor is illustrated in  Figure~\ref{fig:reledobservation} and details are provided in the supplementary material.

\begin{figure}[ht!]
    \includegraphics[width=1\textwidth]{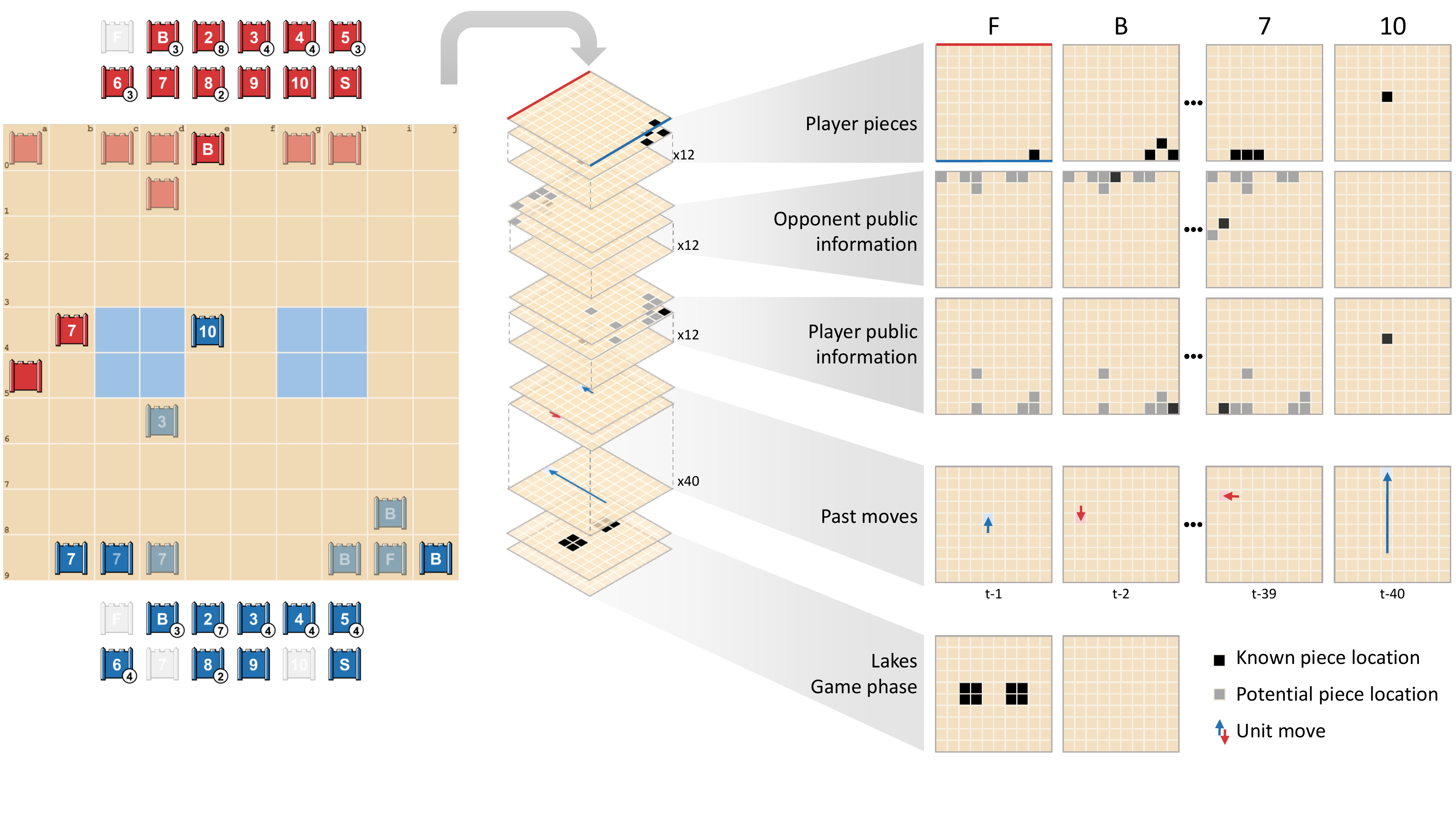}
    \centering
    \caption{The input of the neural network is a single tensor encoding the position of pieces, the currently known information of both opponent and own pieces (whether a piece moved or was  revealed), a limited move history and the position of the lakes.}
    \label{fig:reledobservation}
\end{figure}

\paragraph{The \reled loop:} 
Given a trajectory, the reward transform used at turn $t$ is $r^i_{t,\pi_{m,\textrm{reg}}}(a, \pi) = r^i_t - \eta \log(\frac{\pi(a|o_t)}{\pi_{m, \textrm{reg}}(a|o_t)})$, if $i = \psi_t$ and $r^i_t + \eta \log(\frac{\pi(a|o_t)}{\pi_{m, \textrm{reg}}(a|o_t)})$ if $i \neq \psi_t$ , starting at the initial policy $\pi_{m=0,\textrm{reg}}$.

The \emph{dynamics} step of \agent is composed of two parts, the first part estimates the value function which is done through an adaptation of the $v$-trace estimator~\cite{espeholt2018impala} to the two-player imperfect information case, and the second part learns the policy through the Neural Replicator Dynamics (NeuRD) update~\cite{hennes2020neural} using an estimate of the state action value based on the $v$-trace estimator. These parts are detailed in the supplementary material.

After a fixed number of learning steps, an approximate fixed point policy $\pi_{m,\textrm{fix}}$ is obtained, which is then used as the next regularisation policy: $\pi_{m+1,\textrm{reg}}=\pi_{m,\textrm{fix}}$. The three steps are repeated using a smooth transition from the reward transformation of step $m$ to the one of step $m+1$.


\paragraph{Fine-tuning :} Directly learning with the above-described method leads to convergence to an empirically satisfying solution, which however is slightly distorted by low-probability mistakes. Those mistakes appear because the $\textrm{softmax}$ projection used to compute the policy from the logits assigns a non-zero probability to every action. In order to alleviate this issue we fine-tune during training by performing additional thresholding and discretization to the action probabilities. The supplementary material provides more details on this and also describes a few additional techniques applied at test-time to remove any remaining obvious mistakes from the policy.

\section{Results}
In this section we present an overview of the evaluation results of \agent against both human expert players and current state-of-the-art Stratego bots. For the former we have worked with the Gravon platform, a well-known online games server popular among Stratego players. For the latter we have tested \agent against eight known AI bots that play Stratego. A detailed analysis is also presented with regard to some of the capabilities of the agent's game-play including deployment, bluffing, and trading off of material vs information.

\subsection{Evaluation on Gravon}

\newcommand{\gravonalltimeranking}{\emph{Classic Stratego ranking}\xspace}
\newcommand{\gravontwentytworanking}{\emph{Classic Stratego challenge ranking 2022}\xspace}

Gravon is an internet platform for human players, offering several online games, including Stratego. It is by far the largest online platform for Stratego, where some of the strongest players compete \cite{gravon_site}.
The Gravon platform uses the same rating system as the International Stratego Federation for the world championship (i.e. the Kleier rating \cite{kleierdotnet}).\footnote{Similar to the Elo rating system, the Kleier rating system models the win probability between two players from the difference in their rating.} Gravon offers two rankings: one all-time \gravonalltimeranking and one \gravontwentytworanking. To be included in these rankings, Gravon imposes some limitations to makes sure players are regularly confronted with opponents of comparable strength.

\agent was evaluated against top human players over the course of two weeks in the beginning of April 2022, resulting in 50 ranked matches. Of these matches, 42 (i.e. 84\%) were won by \agent. In the \gravontwentytworanking this corresponds to a rating of 1799, which resulted in a 3\textsuperscript{rd} place for \agent of all ranked Gravon Stratego players (the top two ratings are 1868 and 1831). In the all-time \gravonalltimeranking this resulted in a rating of 1778 which also puts \agent in the 3\textsuperscript{rd} place of all ranked Gravon Stratego players (the top two ratings are 1876 and 1823). The rating for this leaderboard considers all ranked games going back to the year 2002.

These results confirm that \agent reaches a human expert level in Stratego, only through self-play, without bootstrapping from existing human data. 

\subsection{Evaluation against state-of-the-art Stratego bots}

\agent was also evaluated against several existing Stratego computer programs:
\emph{Probe} was a three-fold winner of the Computer Stratego World Championship (2007, 2008, 2010) \cite{stratego_wikipedia};  \emph{Master of the Flag} won that championship in 2009  \cite{stratego_wikipedia}; \emph{Demon of Ignorance} is an opensource implementation of Stratego with an accompanying AI bot \cite{demon_of_ignorance}; \emph{Asmodeus}, \emph{Celsius}, \emph{Celsius1.1}, \emph{PeternLewis}, and \emph{Vixen} are programs that were submitted in 
an Australian university programming competition in 2012 \cite{ucc_australia2012}, won by \emph{PeternLewis}. 

As shown in Table \ref{table:external_bots}, DeepNash wins the overwhelming majority of games against all of these bots, despite not having been trained against any of them and only being trained using self-play. As such it is not necessarily expected that the residual losses against some of these bots would vanish, even if the exact Nash-equilibrium were reached. For example, in most of the few matches that \agent has lost against \emph{Celsius1.1}, the latter played a high-risk strategy of capturing pieces early on with a high-ranking piece, and as such was trying to get a significant material advantage. Most often this strategy does not work, but occasionally it can lead to a win. 

\begin{table}
\begin{center}
\begin{tabular}{ cc|rrr } 
 \toprule
Opponent  & Number of Games & Wins & Draws & Losses \\
 \midrule
Probe & 30  & 100.0\% & 0.0\% & 0.0\% \\
Master of the Flag & 30  & 100.0\% & 0.0\% & 0.0\% \\ 
Demon of Ignorance & 800 & 97.1\% & 1.8\% & 1.1\% \\ 
Asmodeus & 800 & 99.7\% & 0.0\% & 0.3\% \\ 
Celsius & 800 & 98.2\% & 0.0\% & 1.8\% \\ 
Celsius1.1 & 800 & 97.9\% & 0.0\% & 2.1\% \\ 
PeternLewis & 800 & 99.9\% & 0.0\% & 0.1\% \\ 
Vixen & 800 & 100.0\% & 0.0\% & 0.0\% \\ 
 \bottomrule
\end{tabular}
\end{center}
\caption{Evaluation of \agent against existing Stratego bots. The numbers are reported from \agent's point of view. More games (800) were played against bots which we could run automatically. The same number of matches were played as Red and Blue, except against \emph{Master of the Flag} which only plays as Blue. }
\label{table:external_bots}
\end{table}

\subsection{Illustration of \agent's abilities}
\label{sec:behavior}
\agent's only goal during training is to learn a Nash equilibrium policy and by doing so it learns qualitative behavior one could expect a top player to master. Indeed, the agent is able to generate a wide range of deployments which makes it difficult for a human player to find patterns to exploit by adapting their own deployment. We also show situations where \agent is able to make non-trivial trade-offs between information and material, to execute bluffs and to take gambles when needed. The rest of this section illustrates these behaviors through matches that were played on Gravon. 

For convenience, the behavior is described in a way a human observer might naturally interpret it, including terms like "deception" and "bluffing".

\subsubsection{Piece deployment}
\begin{figure}[ht!]
    
    \vspace{-50pt}
   \begin{subfigure}[t]{0.9\textwidth}\centering

    \includegraphics[width=0.4\textwidth]{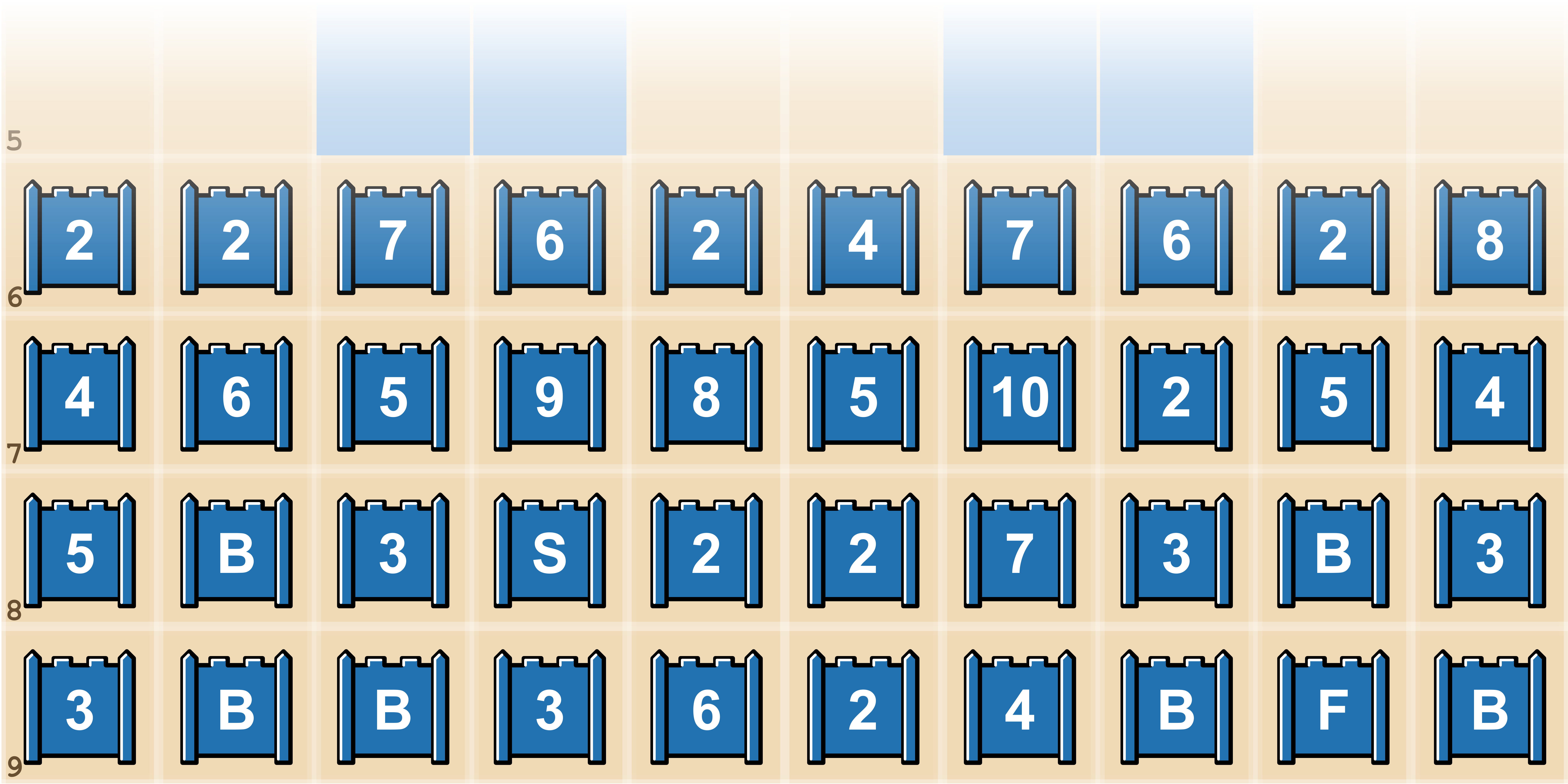} \quad
    \includegraphics[width=0.4\textwidth]{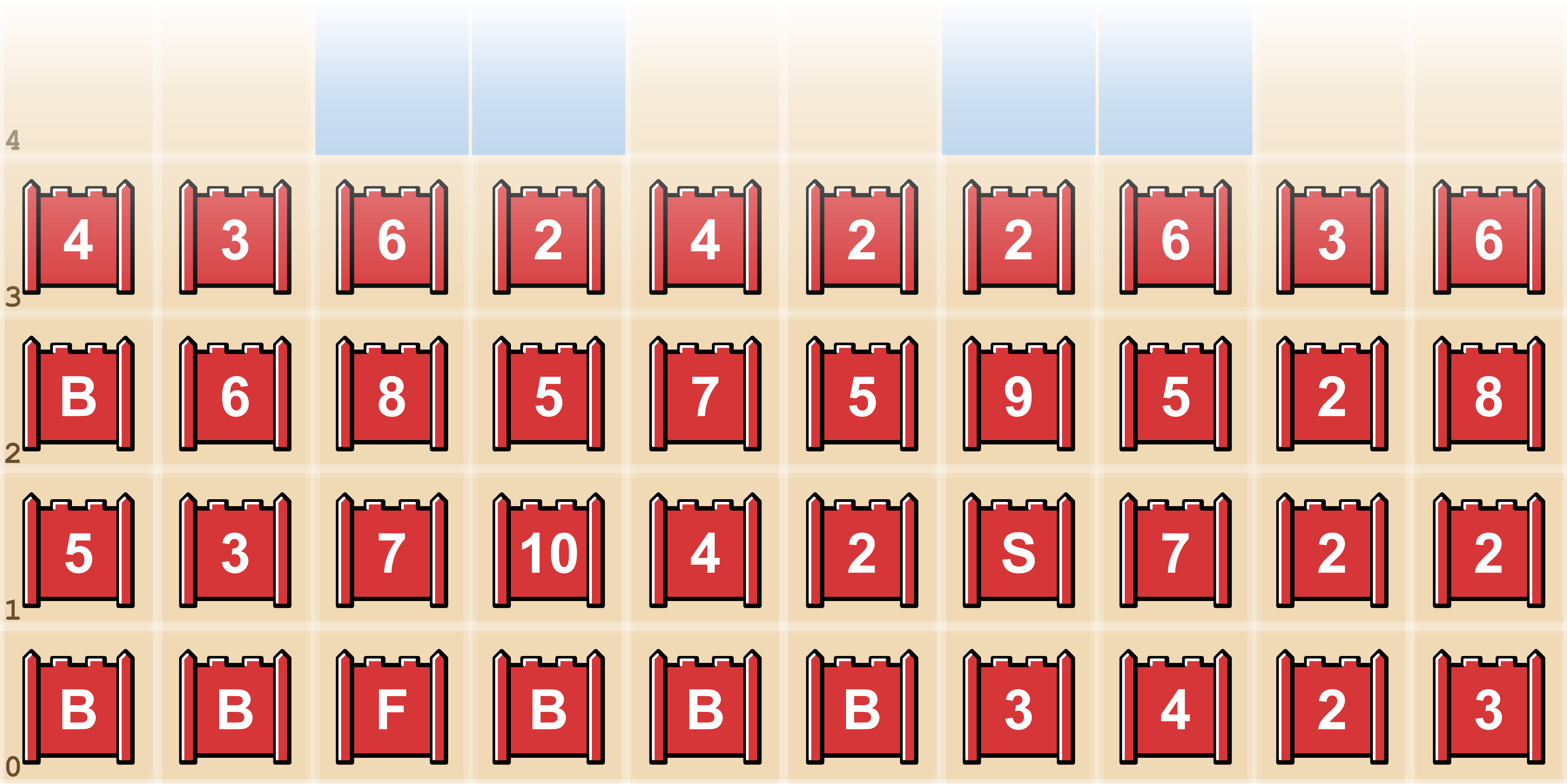}

    \vspace{10pt}

    \includegraphics[width=0.4\textwidth]{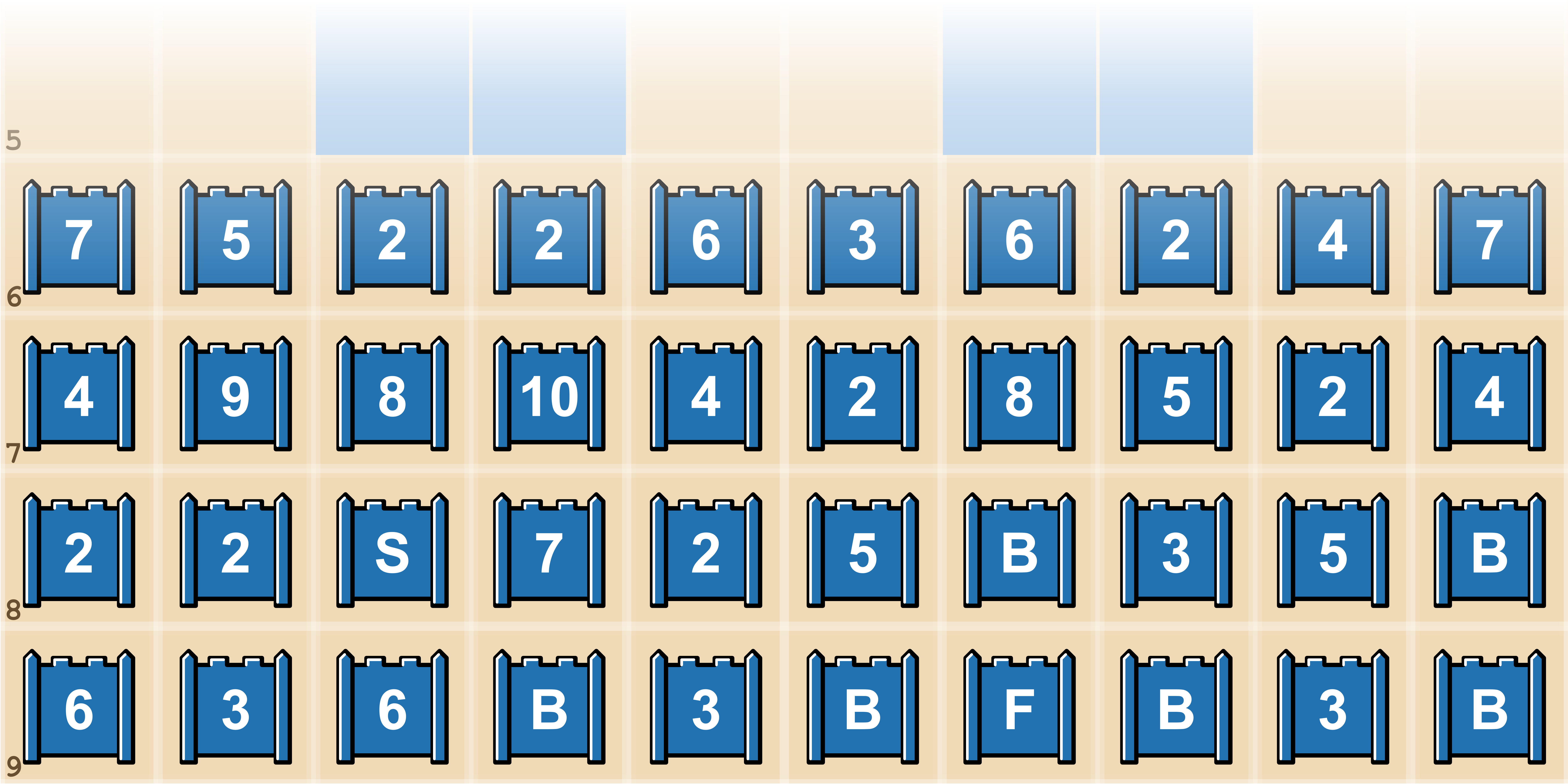} \quad
    \includegraphics[width=0.4\textwidth]{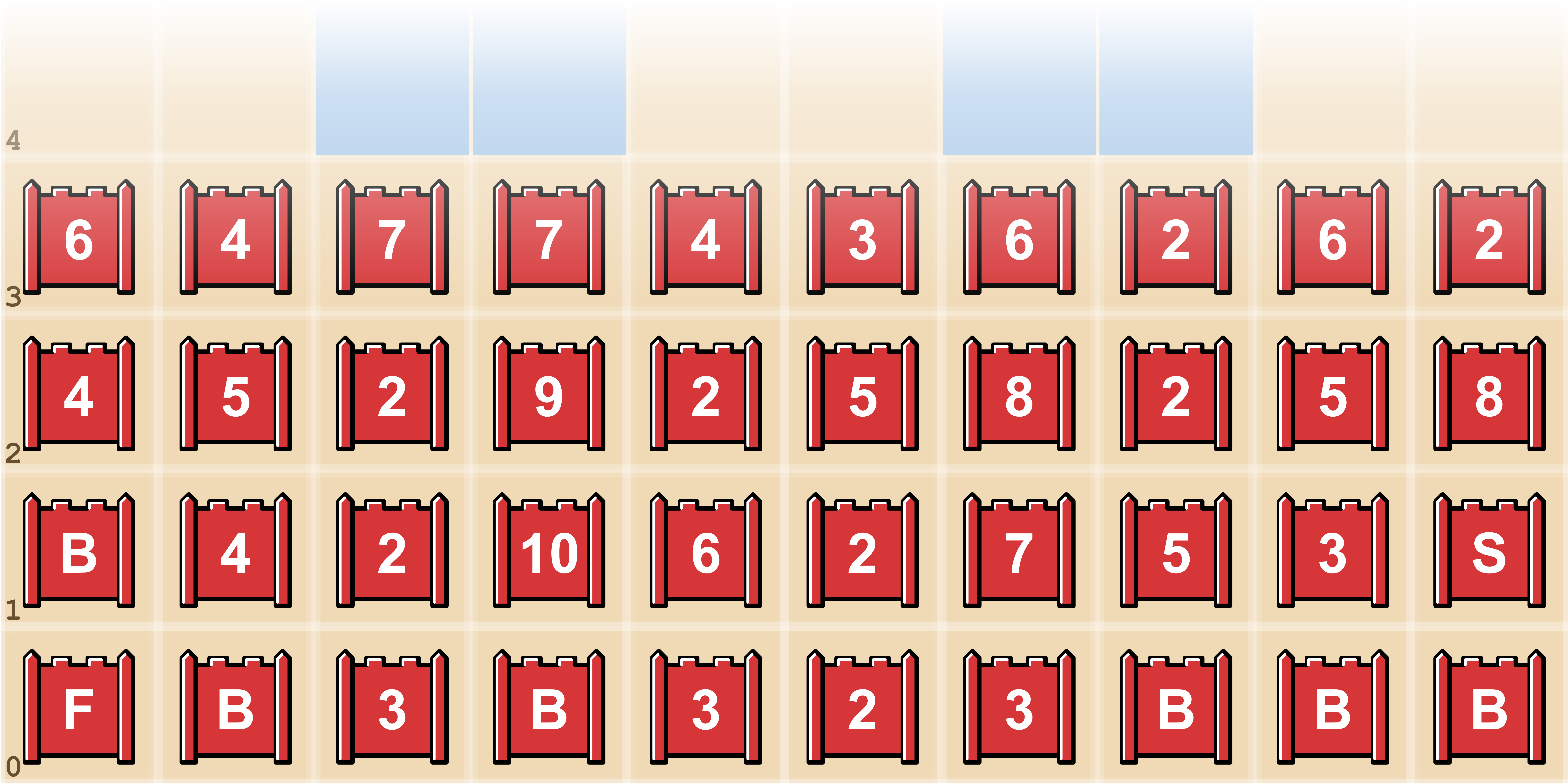}
    \caption{Four example deployments \agent played on Gravon.}
    \label{fig:exampledeployments}
   \vspace{10pt}
   \end{subfigure}
   \centering
   \begin{subfigure}[t]{0.47\textwidth}
        \includegraphics[width=0.9\textwidth]{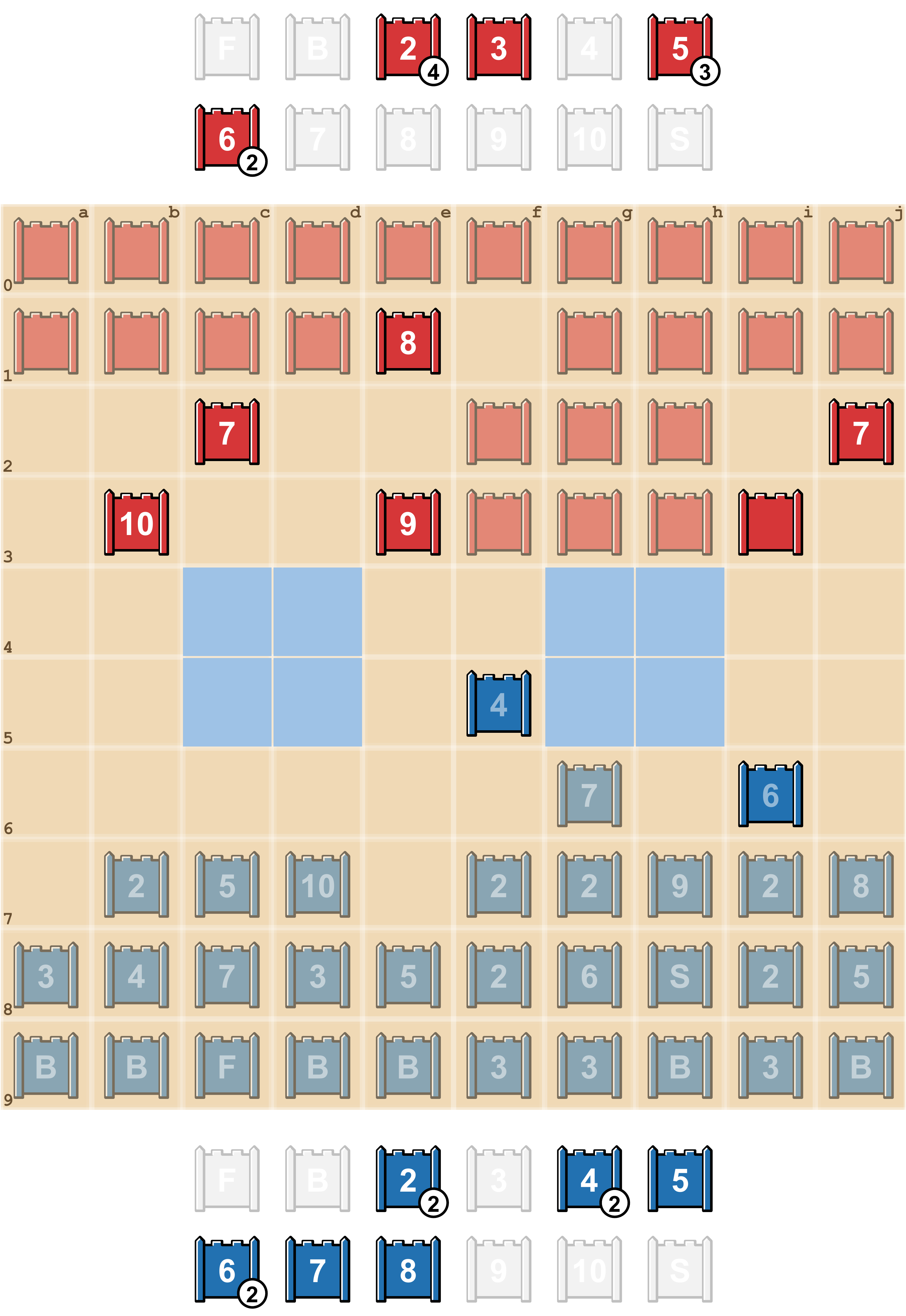}
        \caption{While Blue is behind a 7 and 8, none of its pieces are revealed and only two pieces moved. As a result \agent assesses its chance of winning to be still around 70\%  (Blue indeed won this match).}
        \label{fig:mat_vs_info1}
   \end{subfigure}
   \hfill
   \begin{subfigure}[t]{0.47\textwidth}
        \includegraphics[width=0.9\textwidth]{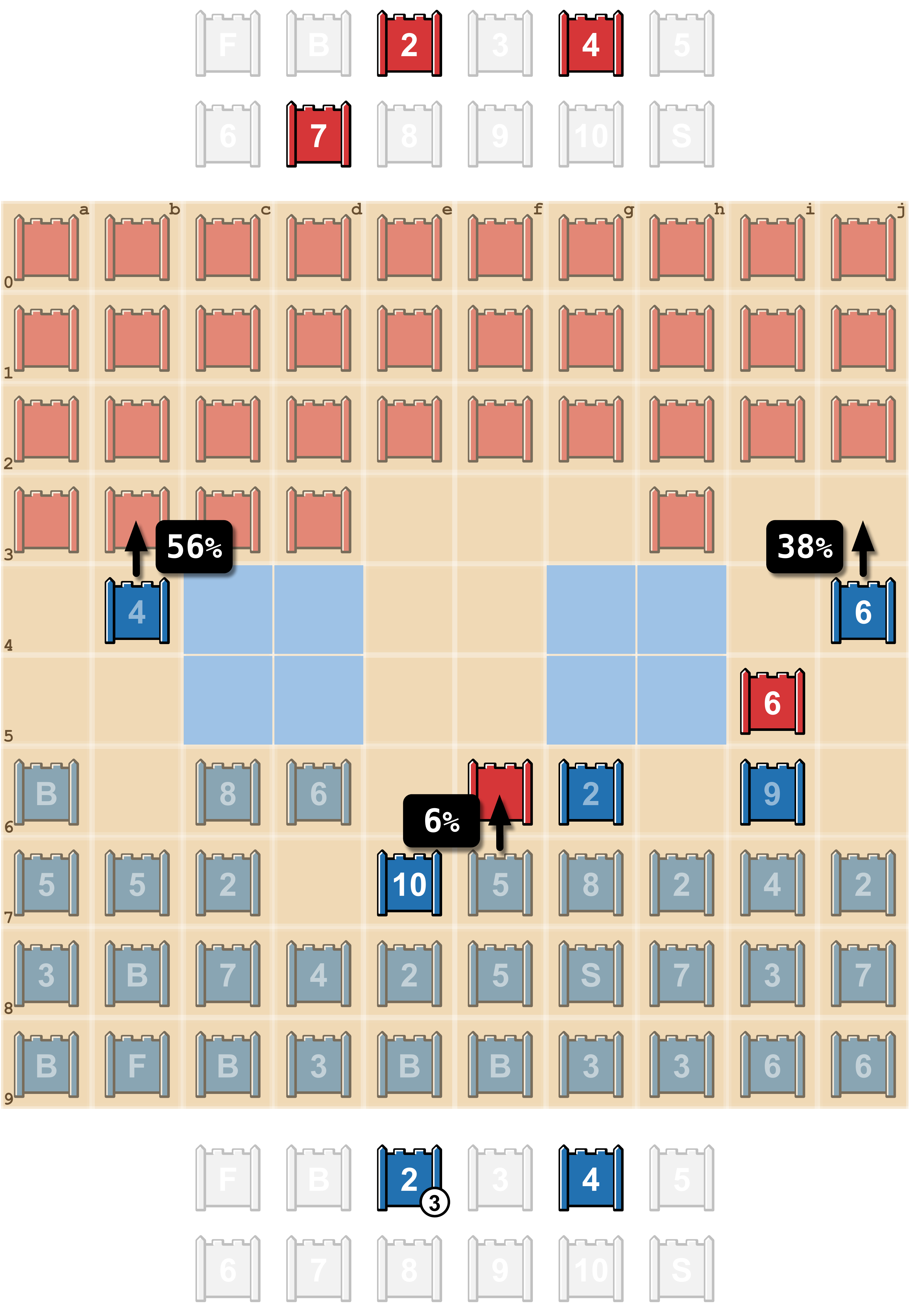}
        \caption{Blue to move. \agent's policy supports three moves at this state, with the indicated probabilities (the move on the right was played in the actual match). While Blue has the opportunity to capture the opponent's 6 with its 9, this move is not considered by \agent, likely because the protection of 9's identity is assessed to be more important than the material gain.}
        \label{fig:mat_vs_info2}
   \end{subfigure}
   \caption{Illustration of \agent's assessment of the relative value of material versus information in two human (red) - \agent(blue) matches. }
   \label{fig:materialvsinfo}
\end{figure}

The imperfect information in Stratego arises during the initial phase of the game where both players place their 40 pieces on the board in a secret configuration. As described above, \agent learns this deployment strategy simultaneously with the regular game-play, and so both strategies co-evolve during learning. Having an unpredictable deployment is important for being unexploitable and indeed \agent is capable of generating billions of unique deployments. 
At the same time, not all possible deployments are equally strong (e.g. putting a Flag in the open on the front row is bad for obvious reasons) and some often recurring deployment patterns by \agent are exemplified in Figure ~\ref{fig:exampledeployments}.

The Flag is almost always put on the back row, and often protected by Bombs. Occasionally, however, \agent will not surround the Flag with Bombs. Experts (e.g. Vincent de Boer, 3-fold World Champion) believe that it is indeed good to occasionally not protect the Flag because this unpredictability makes it harder for the opponent in the end-game. Another pattern observed is that the highest pieces, the 10 and 9, are often deployed on different sides of the board. Additionally, the Spy is quite often located not too far away from the 9 (or 8), which protects it against the opponent's 10. \agent does not often deploy Bombs on the front row, which complies with the behavior seen from strong human players.
The 3's (Miner), which can defuse Bombs, are often placed on the back row, which makes sense because their importance typically increases throughout a game as more opponent Bombs and potential Flag positions get revealed. The eight 2's (Scout) are typically deployed both in the front and more in the back, allowing to scout opponent pieces initially but also in later phases of the game.






%
%
%
%

\subsubsection{Trade-off between information and material}
An important tactic in Stratego is to keep as much information as possible hidden from an opponent in order to gain an advantage. During certain game situations there will be trade-offs to be considered where a player needs to balance the value of capturing an opponent's piece (or even moving a piece), and as such revealing information on their own piece, versus not capturing a piece (or not moving), but keeping the identity of a piece hidden. \agent is able to make such trade-offs in remarkable ways.

Figure \ref{fig:mat_vs_info1} shows a situation where \agent (in blue) is behind in pieces (it lost a 7 and an 8) but is ahead in information as the opponent in red has its 10, 9, an 8 and two of its 7's revealed. Valuing information and material in Stratego is non-trivial a-priori, but the agent has learned a policy through self-play that seems to naturally make this trade-off between information and material. In the above example, \agent is behind in material but knows the identity of many of the opponents' high-ranked pieces. On the contrary, almost all of \agent's remaining pieces have not yet moved and its opponent is left in the blind. The value function ($v=0.403$) credits this information asymmetry as an advantage for \agent (with an expected win rate of around $70\%$) despite having lesser material on the board. This game was won by \agent.

The second example in Figure~\ref{fig:mat_vs_info2} shows a situation where \agent has the opportunity of capturing the opponent's 6 with its 9, but this move is not considered, probably because protecting the identity of the 9 is deemed more important than the material gain. The situation also illustrates the stochasticity of \agent's policy during game-play.

\subsubsection{Deceptive behavior and bluffing}
\begin{figure}
\vspace{-5em}
\captionsetup[subfigure]{slc=off,margin={1cm,0cm}}
  \parbox[t]{0.70\textwidth}{\includegraphics[width=\hsize]{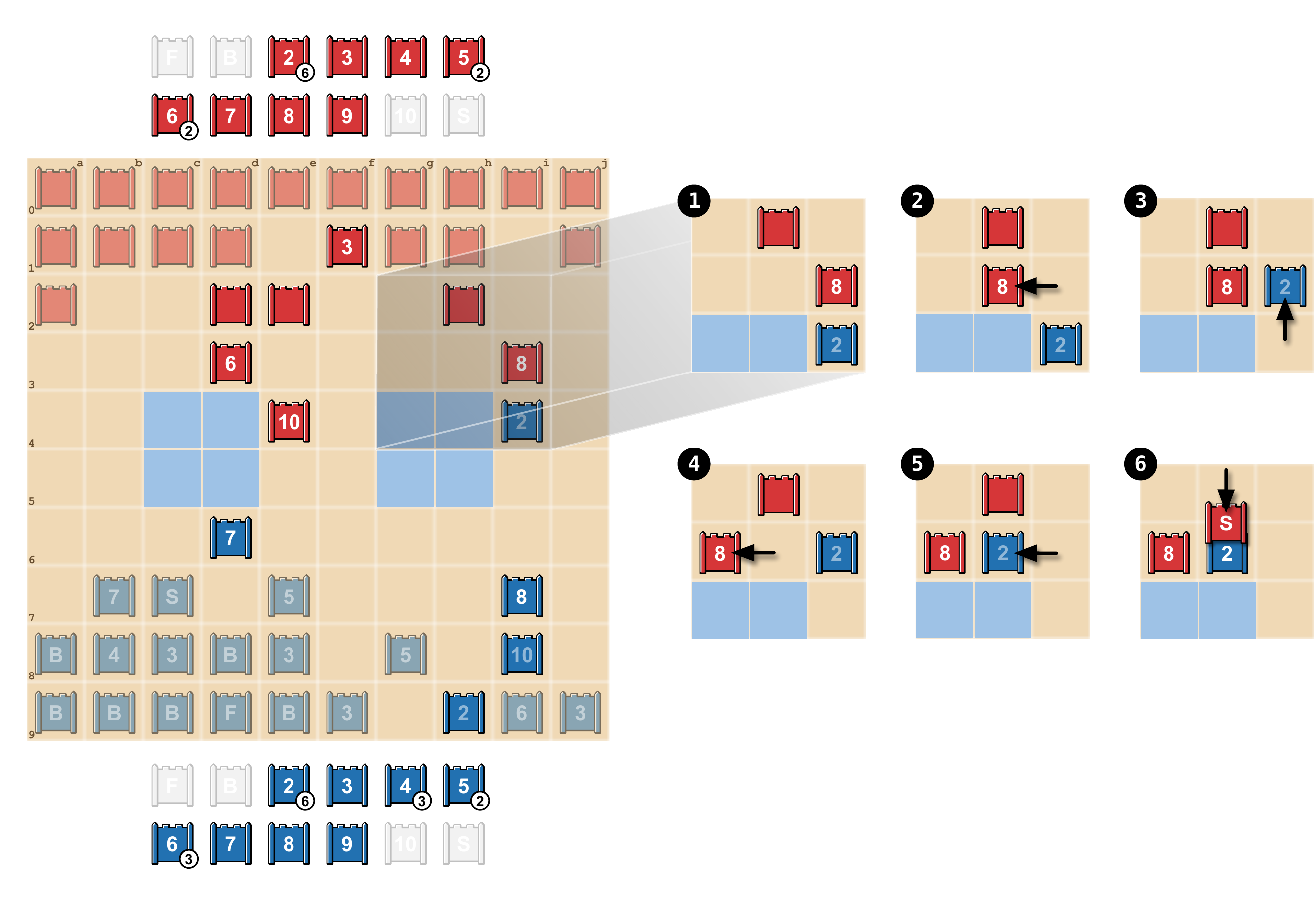}}%
  \parbox[t]{0.30\textwidth}{\vskip -9em \subcaption{\centering Positive bluffing.}\label{fig:bluffing_a}}\\[1ex]
  \vskip -5em
  \parbox[t]{0.30\textwidth}{\vskip -9em \subcaption{\centering Negative bluffing.}\label{fig:bluffing_b}}%
  \parbox[t]{0.70\textwidth}{\includegraphics[width=\hsize]{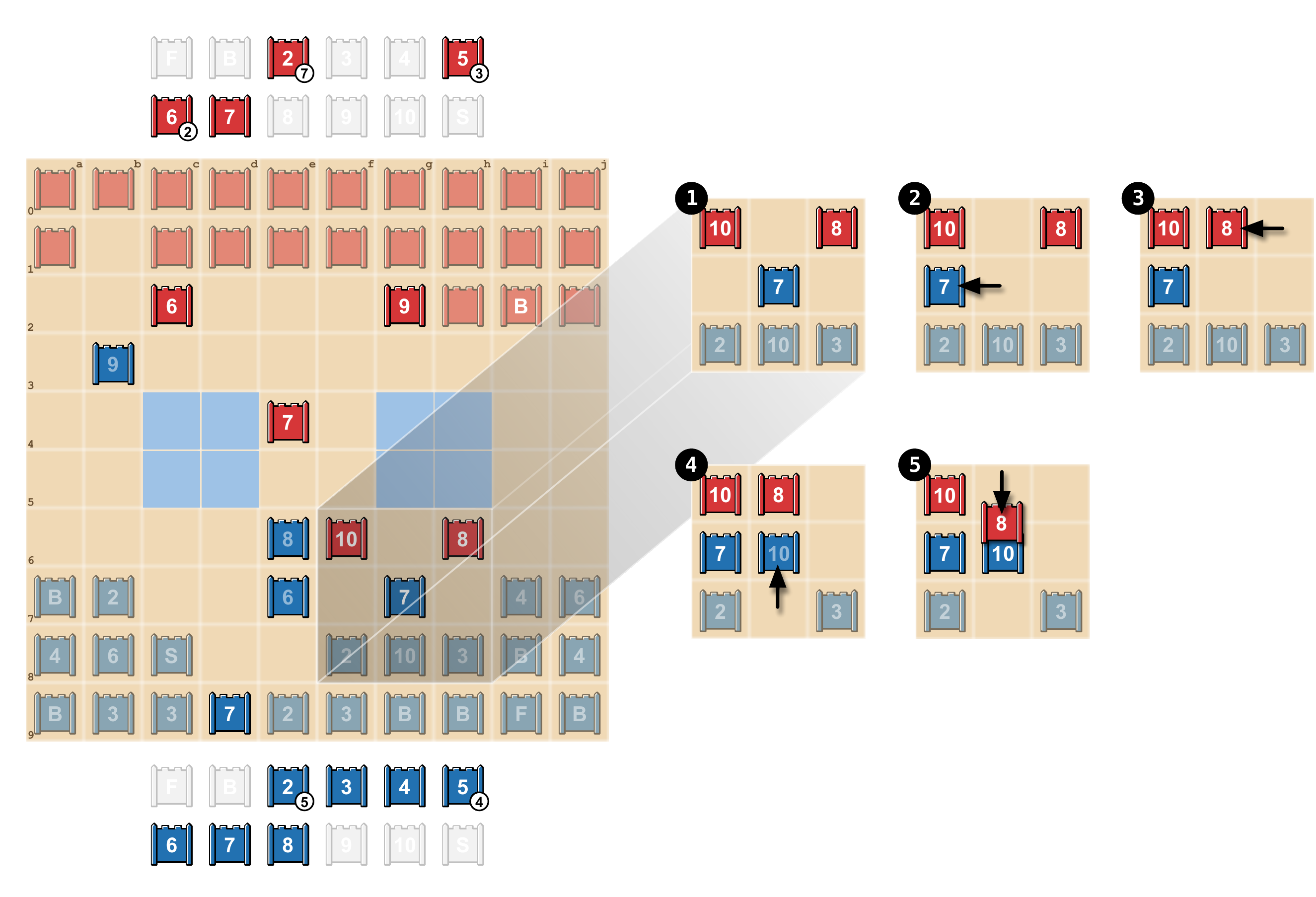}}\\[1ex]
  \parbox[b]{0.82\textwidth}{\includegraphics[width=\hsize]{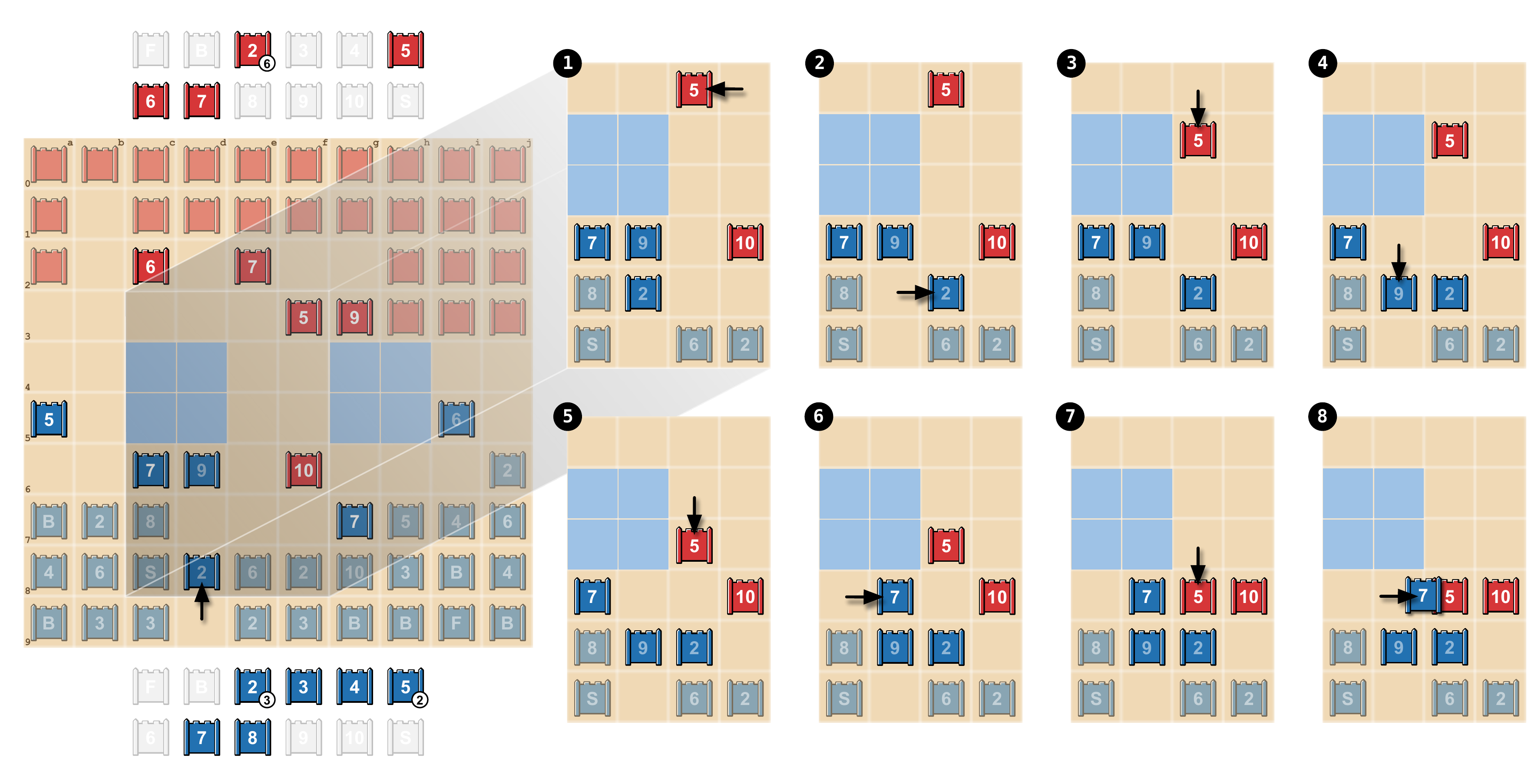}}%
  \parbox[t]{0.2\textwidth}{\vskip -9em \hskip -10em \subcaption{ \agent makes a Scout (2) behave like a Spy and gains material.}\label{fig:bluffing_c}}
  \caption{Illustration of \agent bluffing.}
\end{figure}

In addition to being able to value an asymmetry of information, one can also expect the agent to occasionally bluff in order to deceive its opponent and potentially gain an advantage. The situations shown in Figures~\ref{fig:bluffing_a}, \ref{fig:bluffing_b} and \ref{fig:bluffing_c}  illustrate this ability. In Figure \ref{fig:bluffing_a} we illustrate \emph{positive bluffing}, in which a player pretends a piece to be of a higher value than it actually is. \agent chases the opponent's 8 with an unknown piece, a Scout (2), pretending it to be the 10. The opponent believes this piece has a high chance of being the 10 and guides it next to its Spy (which can capture the 10). In an attempt to capture this piece, however, the opponent loses its Spy to \agent's Scout. 

A second type of bluff, called \emph{negative bluffing}, is shown In  Figure \ref{fig:bluffing_b}, which means that one pretends to be a lower piece as opposed to a positive bluff. Here the movement of the unknown 10 of \agent is interpreted by the opponent as a positive bluff as they try to capture it with a known 8 assuming \agent is moving a lower-ranked piece, potentially the Spy, bringing it closer to the opponent's 10. The opponent instead  encounters \agent's 10 and loses a 8.

A more complex bluff is shown in Figure~\ref{fig:bluffing_c}, where \agent brings its unrevealed Scout (2) close to the opponent's 10, which can be easily interpreted as a Spy. This tactic actually allows Blue to capture Red's 5 with its 7 a few steps later, thereby gaining material but also preventing the 5 from capturing the Scout (2) and revealing it is actually not the Spy.


\section{Conclusion}

In this work, we have introduced a novel method called \agent that learns to play the imperfect information game Stratego from scratch in self-play, up to human expert-level. This model-free learning method combines a deep residual neural network with the game-theoretical \emph{Regularized Nash Dynamics} (\reled) multi-agent learning algorithm, without doing any form of search or explicit opponent modelling. As such, \agent takes an orthogonal approach to state-of-the-art model-based learning methods that have been successfully applied to other complex games such as Go, chess, and imperfect information games such as poker and Scotland Yard, but which, due to their computational toll and the inherent complexity of the Stratego game itself are not applicable to such an elaborate game. \agent learns both the deployment phase of pieces at the start of the game and the actual game-play itself, end-to-end, in one approach.

The core component behind \agent is the at-scale implementation of the \reled algorithm. \agent carries out three essential steps in an iteration of the algorithm: \emph{reward transformation} starting from a random regularised policy to define a modified game, subsequently applying the \emph{replicator dynamics} on this modified game to converge to a fixed point policy, and finally \emph{update} the regularization policy to this new fixed point. Repeatedly applying this three-fold process empirically demonstrates convergence of the learning algorithm to an $\epsilon$-Nash equilibrium in Stratego.

We thoroughly evaluated \agent against eight state-of-the-art Stratego bots and against human-expert Stratego players on the Gravon platform. Against  other AI bots we achieve a minimum win-rate of $97\%$, and in the evaluation against human-expert players we achieve an overall win-rate of $84\%$ on the Gravon platform, which places us in the top-3 rank of both the 2022 and all-times leaderboards. This is a remarkable result that the Stratego community did not believe would have been possible with current techniques, viz. quotes by Thorsten Jungblut (owner of the Gravon platform) and Vincent de Boer, which can be found in the supplemental material. Since June 2021, we have worked on \agent with Vincent de Boer (co-author on this paper), a former three times Stratego world champion, currently ranked 4\textsuperscript{th} on the official world ranking. Vincent has helped evaluating \agent and detecting weaknesses in its consecutive versions, which fed back into agent improvements. 

In conclusion, we believe that \agent can unlock further applications of RL methods in real-world multi-agent problems with astronomical state spaces, characterized by imperfect information, that are currently out of reach for state-of-the-art AI methods to be applied in an end-to-end fashion.

\section*{Acknowledgments}
We would like to thank several colleagues for their feedback on this manuscript and advice during the development of \agent: Laurel Prince, Aja Huang, Martin Schmidt, Michael Bowling, Georg Ostrovski, Martin Riedmiller, Koray Kavukcuoglu, Wojtek Czarnecki, Sukhdeep Singh. We would also like to thank Thorsten Jungblut, the Gravon platform owner, for giving us the opportunity to have access to Gravon, evaluate the \agent agent against human expert players, and for his support, allowing us to use matches played online for illustration in this manuscript.

\newpage
\section*{Related Work}

\subsection*{Regret Minimization and Regularization in Games}
One general class of methods for (approximately) solving games uses self-play of regret-minimizing algorithms.
An algorithm is said to minimize regret if the difference between the average value of the sequence of actions it generates, and that of the best alternate in hindsight, approaches zero over time.
There are different regret measures based off of different sets of alternatives, with the simplest being external regret which considers static alternatives~\cite{hannan1957approximation}.
That is, in a repeated environment with some set of actions $A$ with value $v_t(a)$ for action $a$ at time $t \in \mathbb{N}$ and a sequence of policies $\pi_t$, the external regret is $R_T := \max_{a \in A} \sum_{t=1}^T (v_t(a) - \pi_t \cdot v_t)$.
An external-regret minimizing algorithm, like the regret-matching algorithm~\cite{hart2000simple}, often has a guarantee of the form $R_T \le k\sqrt{T}$ for some constant $k$, which implies $\lim_{T \rightarrow \infty} \frac{R_T}{T}  = 0$.
In the context of zero-sum two-player games, minimizing regret is useful because it is well known that if two regret-minimizing algorithms play against each other, their average policies $1/T \sum_t \pi_t$ approach the set of Nash equilibrium strategy profiles (see for example, \cite{freund1996game}).
However, the action set $A$ grows exponentially in the number of information sets, so game-specific regret-minimizing algorithms are needed for larger extensive form games, with sequential decisions.

The Counterfactual Regret Minimization algorithm~\cite{cfr} applies regret-matching independently at each information set, with a proof that minimizing regret at all information sets also minimizes external regret.
The simplicity and computational efficiency of CFR has led to many variants, like the sampling variant MCCFR~\cite{lanctot2009monte}, Pure CFR~\cite{gibson2014}, CFR+~\cite{tammelin2015solving}, and Discounted CFR~\cite{brown2019solving}.
However, CFR is a tabular method that stores regret information and action probabilities for all information sets in the game, so is limited to games which can fit in storage.
Using compression, tabular CFR variants have been applied to games with up to $10^{14}$ information sets~\cite{tammelin2015solving}, but are incapable of scaling up over a hundred orders of magnitude to deal with games like Stratego.

Follow the Regularized Leader (FoReL) is another approach to online learning.
The motivation for FoReL comes from examining the behaviour of the simple rule of Follow the Leader, which looks back to find the historically best decision and follows that choice at the current time.
While this is a natural and perhaps intuitive approach to sequential interactions, Follow the Leader generally lacks any interesting guarantee on performance.
To get such a guarantee, there are a number of different intuitions which lead to the idea of adding a bonus (or penalty depending on the point of view) to the historically-observed values, like dampening the response or adding a strongly-convex term to the optimisation.
The resulting approach, of following the historically-best decision when including an additional regularization term, describes the class of follow the regularized leader (FoReL) algorithms.
Without game-specific approaches or function approximation, FoReL has the same limitations as regret-matching: the direct application would be on an exponentially large space of actions, and it is a tabular method that tracks historical values for all actions.

\subsection*{Reinforcement Learning in Two-Player Zero-Sum Games}
Reinforcement learning methods have been applied to two-player zero-sum games going back to TD-Gammon \cite{tesauro1995temporal}. However until recently, most successful methods were limited to the perfect information case \cite{tesauro1995temporal, buro1998simple, schaeffer2001temporal, silver2017mastering, silver2018general}. New deep RL methods based on regret minimization, best responses, and policy gradients have shown success in imperfect-information games such as poker.  

The first category of deep RL algorithms are based on regret minimization techniques such as counterfactual regret minimization (CFR) \cite{cfr}. Deep CFR \cite{deep_cfr} approximates CFR by training a regret network on a buffer of counterfactual values. However, Deep CFR uses external sampling, which may be impractical for games with a large branching factor such as Stratego and Barrage Stratego. DREAM \cite{steinberger2020dream} and ARMAC \cite{gruslys2020advantage} are model-free regret-based deep learning approaches that can take advantage of outcome sampling and can therefore scale to large games. However, they rely on importance sampling terms to remain unbiased, and this importance weight might become very large in games with a long horizon such as Stratego. Furthermore, all these techniques when generalized to using neural networks require generating an average strategy which is either memory heavy (as one needs to store all strategies all iterations to get an exact average) or error prone when using approximation (DeepCFR for instance use a supervised learning step to approximate the average).

The second category of deep RL algorithms for two-player zero-sum games include best-response techniques. Generally, best response techniques iteratively train a best response via reinforcement learning every iteration. Neural Fictitious Self Play (NFSP) \cite{nfsp} approximates extensive-form fictitious play by progressively training a best response against an average of all past policies using off-policy reinforcement learning. The average policy is a neural network that is trained to imitate the average of the past best responses. Policy Space Response Oracles (PSRO) \cite{psro} iteratively adds a reinforcement learning best response to a population of policies, one for each player. The best response is computed by training against a meta-distribution over the current population policies. This meta-distribution is computed by finding a Nash equilibrium of the empirical game matrix formed by considering each policy as a pure strategy in a normal form game. AlphaStar \cite{alphastar} beat top humans at Starcraft using a method inspired by PSRO, where several agents were training at the same time against a dynamically-updated meta-distribution over all policies. Similarly, OpenAI Five \cite{dota} beat top humans at Dota using a mixture of self-play and a dynamically-updated meta-distribution over past policies. Finally, a similar population-based method combined with population-based hyperparameter optimization has achieved human-level performance on Capture the Flag \cite{pbt}. Despite these successes, best response techniques remain memory-intensive because they potentially require an exponential (in the number of distinct states) number of different policies to represent an optimal policy, which, combined with the time required to compute a best-response, also makes them very slow.

The third category of deep RL algorithms, and where this work falls under, is policy gradient methods. Regret Policy Gradient (RPG) \cite{qpg} approximates CFR via a weighted policy gradient, but is not proven to converge to a Nash equilibrium. Neural Replicator Dynamics (NeuRD) \cite{Neurd} approximates Replicator Dynamics with a policy gradient and is proven to converge to a Nash equilibrium in the time average. Prior to this work, neither of these algorithms have been applied to large-scale domains, or have demonstrated human-level performance; this work uses NeuRD combined with the regularization idea laid out in~\cite{perolat2020poincar} to converge in the last iterate.

\subsection*{Work on Barrage Stratego}
Recently, a smaller variant of Stratego (Barrage Stratego Barrage) has seen progress from a reinforcement learning perspective. Current Barrage Stratego bots are based on imperfect information tree search and are unable to beat intermediate-level human players \cite{stratego_bot, stratego_bots}. Pipeline PSRO \cite{mcaleer2020pipeline} was able to beat these handcrafted bots in Barrage Stratego (by at most 81\% win-rate), but didn't show results against top human players. 

\section*{Methods: additional information}

\subsection*{Details on \reled in Two-Player Zero-Sum Normal Form Games}\label{secapp:reled}
We now concisely describe the above process more formally in the context of NFGs. In a two-player zero-sum normal-form game, player $1$ and $2$ simultaneously play  actions $a^1 \in A^1$ and $a^2 \in A^2$ with policies $\pi^1(a^1)$ and $\pi^2(a^2)$. As a result, player $1$ receives a reward $r^1(a^1, a^2)$ and player $2$ will receive the opposite reward $r^2(a^1, a^2) = -r^1(a^1, a^2)$ (due to the zero-sum nature of the game). The \textit{reward transformation} step of \reled is defined based on a regularization policy $\pi_{\textrm{reg}} = (\pi^1_{\textrm{reg}}, \pi^2_{\textrm{reg}})$ which modifies the reward as follows: $r^i(\pi^i, \pi^{-i},a^i, a^{-i}) = r^i(a^i, a^{-i}) - \eta \log(\frac{\pi^i(a^i)}{\pi^i_{\textrm{reg}}(a^i)}) + \eta \log(\frac{\pi^{-i}(a^{-i})}{\pi^{-i}_{\textrm{reg}}(a^{-i})})$, with $\eta > 0$ a regularization parameter.\footnote{We follow the convention of denoting the opponent of player $i$ as $-i$. } The initial regularization policy can be chosen arbitrary (as long as all actions have a non-zero probability).

The \textit{dynamics} step of \reled determines a new policy $\pi_{\textrm{fix}} = (\pi^1_{\textrm{fix}}, \pi^2_{\textrm{fix}})$ derived from convergence to a fixed point of the modified game by applying the replicator dynamics~\cite{smith,smith2,Zeeman80,Zeeman81}. Replicator dynamics are a learning process that are also known as an instance of Follow the Regularized Leader~\cite{McMahan11a}, which are defined as follows: 
$$  \frac{d}{d\tau}\pi^i_\tau(a^i) = \pi^i_\tau(a^i)[Q^i_{\pi_\tau}(a^i)-\sum\limits_{b^i}\pi^i_\tau(b^i)Q^i_{\pi_\tau}(b^i)],\textrm{with} \quad Q^i_{\pi_\tau}(a^i) = \mathbb{E}_{a^{-i}\sim \pi^{-i}_\tau}\left[r^i(\pi^i_\tau, \pi^{-i}_\tau,a^i, a^{-i})\right]$$
The replicator dynamics are a descriptive learning process from evolutionary game theory that aims to reinforce the probability of the actions with a high fitness $Q^i_{\pi_\tau}(a^i)$ ($Q$ standing for \emph{Q}uality of an action), and decreasing the probability of the actions with low fitness \cite{Zeeman80,Zeeman81,smith,smith2,BloembergenTHK15}. As such it measures the expected payoff for an action vs. that of the average of all actions. If an action performs better than average its probability will increase, otherwise decreasing. This dynamical system has a fixed point $\pi_{\textrm{fix}}$ and convergence to it is guaranteed through the Lyapunov function  
$H_{\pi_{\textrm{fix}}}(\pi) = \sum \limits_{i=1}^{2} \sum \limits_{a^i \in A^i} \pi^i_{\textrm{fix}}(a^i)\log\left(\frac{\pi^i_{\textrm{fix}}(a^i)}{\pi^i(a^i)}\right)$. In this case it is a strong Lyapunov function of the system (in fact $\frac{d}{d\tau}H_{\pi_{\textrm{fix}}}(\pi_\tau)\leq \ -\eta H_{\pi_{\textrm{fix}}}(\pi_\tau)$) which means that the distance to the fixed point will decrease exponentially to $0$.

Finally, in the \textit{update} step of \reled we use the previously-obtained fixed point as the regularization policy of the next iteration. So the whole process can be described as follows using an extra iteration index: one starts with an arbitrary regularisation policy $\pi_{0,\textrm{reg}}$. Then given any such regularization policy $\pi_{n,\textrm{reg}}$, we compute the fixed point $\pi_{n,\textrm{fix}}$ under the replicator dynamics of the game with transformed reward. Finally we choose $\pi_{n+1,\textrm{reg}} = \pi_{n,\textrm{fix}}$ and start the next iteration. This process generates a sequence of fixed points $n \rightarrow \pi_{n,\textrm{fix}}$, which is known to converge to the Nash equilibrium\footnote{For simplicity we assume the Nash equilibrium is unique.}, $\pi_{\textrm{nash}}$, of the original game as the sequence of distances to this Nash equilibrium $n \rightarrow \sum \limits_{i=1}^2 \text{KL}(\pi^i_{\textrm{nash}},\pi^i_{n,\textrm{fix}})$ can be proven~\cite{perolat2020poincar} to decrease to $0$ where $\text{KL}$ is the Kullback–Leibler divergence.

The regularization parameter $\eta$ is fixed throughout this process. Its value has two effects on the dynamics step of \reled: on the one hand a higher value gives more stable and faster convergence to the fixed point. On the other hand a higher $\eta$ results in a fixed point that is more biased towards the regularization policy, which means one might need more overall iterations to approach the Nash equilibrium sufficiently close. 

\subsection*{DeepNash: \reled at Scale}

As a reminder, \agent consist of three components: (1) the core component is \reled, the model-free RL algorithm implemented using a deep Neural Network, (2) fine-tuning of the learnt policy to reduce the residual probabilities of taking highly improbable actions and, (3) test-time improvements to avoid remaining obvious mistakes.  

We first lay out some essential background information on imperfect information games necessary to understand how \reled is scaled with a deep neural network. Then we continue to unpack the three algorithmic steps of \reled and detail how they are implemented in the neural architecture. 

\subsubsection*{Imperfect Information Games}
\label{sec:iignotation}
In a two-player Imperfect Information Game, each player (player $1$ and player $2$) play in turn starting from an initial history $h_{\textrm{init}}$. The set $\mathcal{H}$ is the set of all histories and $\mathcal{A}$ is the set of all possible actions. In each history $h\in \mathcal{H}$, the current player takes an action $a\in \mathcal{A}$. As a result of this action $a$, the player $i$ receives a reward $r^i(h, a)\in \mathbb{R}$ and the history is updated to $h'=ha$ (the concatenation of both history $h$ and action $a$). For any given history $h$ the player's turn is noted $\psi(h) \in \{1, 2\}$. The information state $x(h)$ is the set of all histories $h' \in x(h)$, which are indistinguishable from $h$ from player $\psi(h)$'s point of view. We consider the information set to be of perfect recall (there is as much information in the information set $x(h)$ as in the sequence of information sets that were seen by the player until history $h$).
Each player's goal is to produce a policy $\pi^i(a|o)$ where $o$ is the observation given to the player at history $h$ (we overload the notation and also write $o(h)$ the observation function). As described above we will also consider policy dependent rewards $r^i(\pi, h, a)$ used by \reled.

For a given joint policy $\pi = (\pi^1, \pi^2)$ the value of an history $h$ for player $i$ is :
$$v^i_\pi(h) = \sum \limits_{a \in \mathcal{A}} \pi^{\psi(h)}(o(h))\left[r^i(\pi, h, a) + v^i_\pi(ha)\right]$$. The $Q$-function of a policy $\pi$ is defined as $Q^i_\pi(h, a) = r^i(\pi, h, a) + v^i_\pi(ha)$. This function expresses how good it is to take action $a$ from history $h$. In addition, for a given policy the reach probability of a history $h$, $\rho_\pi(h)$, is defined recursively $\rho_\pi(ha) = \pi^{\psi(h)}\left(a|o(h)\right)\rho_\pi(h)$, expressing how probably it is for history $h$ to occur.
The value given an observation is defined as $v^i_\pi(o) = \frac{\sum \limits_{h\in o}\rho_\pi(h) v^i_\pi(h)}{\sum \limits_{h\in o}\rho_\pi(h)}$.

In a self-play, model free reinforcement learning setting, the agent plays against itself using the policy $\pi$ starting from $h_0 = h_{init}$. at each time $t$ the player samples an action $a_t$ according to $\pi(.|o_t)$ (with $o_t = o(h_t)$ and $r^i_t=r^i(h_t, a_t)$) and the state becomes $h_{t+1} = h_{t} a_t$ at time $t+1$. The following trajectories are collected $\mathrm{T} = [(o_t, a_t, (r^1_t, r^2_t), \mu_t(.)=\pi(.|o_t)), \psi_t=\psi(h_t)]_{0\leq t <t_{\textrm{max}}}$ and are the only information we will use during training. We also write $t_{\textrm{effective}}$ the effective length of the trajectory if it ended up finishing before $t_{\textrm{max}}$.

\subsubsection*{Model-free Reinforcement Learning with Regularized Nash Dynamics}
Again, as a reminder, \agent is essentially the \reled algorithm at scale using a deep neural network. As in NFGs, it is done in 3 steps: (1) the \emph{reward transformation step}, (2) the \emph{dynamics step} which is used to empirically converge to a fixed point, and (3) the \emph{update step} in which the algorithm updates the policy used to define the regularization function.

\reled's learning update generates a sequence of policy and value parameter $\theta_n$ indexed by $n$, and of target parameter $\theta_{n, \textrm{target}}$ (the policies are written $\pi_{\theta_n}$ and $\pi_{\theta_{n, \textrm{target}}}$ and the values are written $v_{\theta_n}$ and $v_{\theta_{n, \textrm{target}}}$). The policy $\pi_\theta$ is defined as $\pi_\theta(.|o) = \frac{\exp(l_\theta(.|o))}{\sum\limits_b \exp(l_\theta(b|o))}$ ($l_\theta(.|o)$ is called the logit of the policy).

\paragraph{\textit{Transformation} of the reward:} 
The reward transformation\footnote{The reward transformation used to train \agent is based on the theory developed in~\cite{perolat2020poincar} for imperfect information zero-sum games. It describes both the convergence properties and the Lyapunov functions used in sequential imperfect information games. Here we demonstrate how to practically scale these principles to learn in zero-sum games at scale.}  of \reled is done over an interval of size $\Delta_{m}$. It is based on the policy dependent reward $r^i_{\pi_{m,\textrm{reg}}}(\pi, h, a) = r^i(h,a) + (1-2\times \mathbf{1}_{i = \psi(h)})\eta \log(\frac{\pi(a|o(h))}{\pi_{m, \textrm{reg}}(a|o(h))})$ starting with $\pi_{-1,\textrm{reg}}=\pi_{0,\textrm{reg}}$. At each step $n\in [0,\Delta_m]$ the reward transformation is an interpolation between $r^i_{\pi_{m,\textrm{reg}}}$ and $r^i_{\pi_{m-1,\textrm{reg}}}$. The reward at step $n$ is defined as : $r^i_{\textrm{reg},n}(\pi, h, a) = \alpha_n r^i_{\pi_{m,\textrm{reg}}}(\pi, h, a) + (1-\alpha_n)r^i_{\pi_{m-1,\textrm{reg}}}(\pi, h, a)$ (with $\alpha_n = \min(1,2\times\frac{n}{\Delta_{m}})$) in order to smooth the transition between a regularization policy to another (also $\pi_{m, n, \textrm{reg}} = \alpha_n \pi_{m,\textrm{reg}} + (1-\alpha_n)\pi_{m-1,\textrm{reg}}$). We write $\mathrm{T}_n$ for the trajectory with the transformed reward at step $n$.

The \emph{Dynamics} step at scale of the \reled algorithm is composed of two parts, one concerns the estimation of the value function, and the second part concerns learning update of the $Q$-function and of the policy.

\paragraph{\textit{Dynamics} : estimators for the value function.}
The fixed point $\pi_{m,\textrm{fix}}$ associated to the regularization policy $\pi_{m,\textrm{reg}}$ is learnt over $\Delta_{m}$ steps using two learning updates: (1) an update to learn a value function and generate an estimate of the $Q$-function and (2) an update to learn a policy from the estimated $Q$-function.

It is challenging in Stratego to generate a good estimate of a $Q$-function as the action space is vast with 3600 possible actions at every step, even if not all are legal actions (e.g. a trapped piece that cannot move). The estimators at time $t$ and at learning step $n$ of the value function $\hat v^i_{t,n}$ and of the $Q$-function $\hat Q^i_{t,n}$ for player $i$ adapt the $v$-trace estimator~\cite{espeholt2018impala} to the two player case. It uses information on the future steps in order to create a low variance and low bias estimator of the value and the $Q$-function of the policy $\pi_{\theta_n}$, even if the trajectory was generated with a different policy.

This recursive backward process takes as an input a joint policy $\pi$, a joint value $v$, and a trajectory $\mathrm{T}$ and output $\hat v^1_t$, $\hat v^2_t$, $\hat Q^1_t$ and $\hat Q^2_t$ (we write $\hat v^1_t, \hat v^2_t, \hat Q^1_t, \hat Q^2_t = \Upsilon(\mathrm{T}, \pi, v)$). Given a trajectory $\mathrm{T} = [(o_t, a_t, (r^1_t, r^2_t), \mu_t(.), \psi_t=\psi(h_t)]_{0\leq t <t_{\textrm{max}}}$ (with transformed rewards) the backward update is defined as follow for all player in $[1, 2]$ starting from $t=t_{\textrm{effective}}$ (and $\hat v^i_{t_{\textrm{effective}}+1} =0$, $V^i_{\textrm{next},t_{\textrm{effective}}+1}=0$, $\hat r^i_{t_{\textrm{effective}}+1}=0$ and $\xi_{t_{\textrm{effective}}+1}=1$) :\newline
if $i \neq \psi_t$:
\begin{align}
\hat v^i_t = \hat v^i_{t+1}, \;
V^i_{\textrm{next},t} = V^i_{\textrm{next},t+1}, \;
\hat r^i_t = r^i_t + \frac{\pi(a_t|o_t)}{\mu_t(a_t)} \hat r^i_{t+1}, \;
\xi_t = \frac{\pi(a_t|o_t)}{\mu_t(a_t)}\xi_{t+1}
\end{align}
if $i = \psi_t$:
\begin{align}
&\hat v^i_t = v(o_{t}) + \delta_t V^i + c_t (v^i_{t+1}-V^i_{\textrm{next},t+1}), \;
V^i_{\textrm{next},t} = v(o_{t}) \\
&\delta_t V^i = \rho_t ( r_t + \frac{\pi(a_t|o_t)}{\mu_t(a_t)} \hat r^i_{t+1} + V^i_{\textrm{next},t+1} - v(o_{t})), \;
\hat r^i_t = 0, \;
\xi_t = 1\\
&\rho_t = \min(\bar \rho , \frac{\pi(a_t|o_t)}{\mu_t(a_t)}\xi_{t+1}), \;
c_t = \min(\bar c , \frac{\pi(a_t|o_t)}{\mu_t(a_t)}\xi_{t+1})\\
&\hat Q^i_t (a) = -\eta\log(\frac{\pi_{\theta_n}(a|o_t)}{\pi_{m, n, \textrm{reg}}(a|o_t)})\\
&\qquad + \frac{\mathbf{1}_{a=a_t}}{\mu_t(a_t)}\left(r^i_t + \eta\log(\frac{\pi_{\theta_n}(a|o_t)}{\pi_{m, n, \textrm{reg}}(a|o_t)}) + \frac{\pi(a_t|o_t)}{\mu_t(a_t)} (\hat r^i_{t+1} + \hat v^i_{t+1})-v(o_{t})\right)+v(o_{t})\nonumber
\end{align}
These estimator are computed over the full trajectory from the end to the beginning without any bootstrapping. This ensures that these estimates have a minimum bias.
\paragraph{\textit{Dynamics} : learning update of the $Q$-function and of the policy.} First we use $\hat v^1_{t,n}, \hat v^2_{t,n}, \hat Q^1_{t,n}, \hat Q^2_{t,n} = \Upsilon(\mathrm{T}_n,  \pi_{\theta_{n, \textrm{target}}}, v_{\theta_{n, \textrm{target}}})$ as estimate to compute the value and policy update.

The value is learned through a regression loss written $l_{\textrm{critic}}(\theta) = \sum \limits_{i=1}^2 \frac{1}{t_{\textrm{effective}}}\sum \limits_{t=0}^{t_{\textrm{effective}}}\mathbf{1}_{i=\Psi_t}\|v_{\theta}(o_t)-\hat v^i_t\|$ and the policy is updated through the Neural Replicator Dynamics (NeuRD) loss~\cite{hennes2020neural} and the update direction is defined as:
$$\Lambda_n = -\left[\textrm{lr}_n \nabla l_{\textrm{critic}}(\theta_n) + \sum \limits_{i=1}^2 \frac{1}{t_{\textrm{effective}}}\sum \limits_{t=0}^{t_{\textrm{effective}}} \sum\limits_a \hat\nabla\theta(l_{\theta_n}(a, o_t)\textrm{Clip}\left(Q^{\psi_t}_{t,n}(a, o_t), c_{\textrm{clip NeuRD}}\right), \textrm{lr}_n, \beta)\right]$$
with $\hat\nabla\theta(z(\theta), \eta, \beta) = \eta \nabla\theta z(\theta)\mathbf{1}_{z(\theta + \eta \nabla\theta z(\theta))\in [-\beta, \beta]}$, $\textrm{Clip}(., c) = \min(\max(., -c), c)$ and finally the parameters are updated through an adam optimizer :
$$\theta_{n+1}=\textrm{Adam}(\theta_n, \textrm{Clip}(\Lambda_n, c_{\textrm{clip gradient}}), b_{1,\textrm{adam}}, b_{2,\textrm{adam}}, \epsilon_{\textrm{adam}})$$
and,
$$\theta_{n+1,\textrm{target}} = \gamma_{\textrm{averaging}}\theta_{n+1}+(1-\gamma_{\textrm{averaging}})\theta_{n,\textrm{target}}\textrm{ with $\gamma_{\textrm{averaging}}\in[0, 1[$}$$

\paragraph{\textit{Updating} the transformed reward and the learning parameters.} After the dynamics steps of the algorithm is completed we define the new fixed point policy as $\pi_{m,\textrm{fix}}=\pi_{\theta_{n=\Delta_m}, \textrm{target}}$. The next regularisation policy is defined as : $\pi_{m+1,\textrm{reg}}=\pi_{m,\textrm{fix}}$ and we go on to the next step ($m+1$) starting from the parameters $\theta_{0}$, $\theta_{0, \textrm{target}}$ and the state of the optimizer being the ones we finished the iteration $m$ with. the new reward transform at step $m+1$ interpolates between $r^i_{\pi_{m+1,\textrm{reg}}}$ and $r^i_{\pi_{m,\textrm{reg}}}$.

\paragraph{Fine-tuning :} Learning only with the previous method is enough to converge to an empirically satisfying solution but limited by low probability mistakes. Those mistakes appear because the $\textrm{softmax}$ projection used to compute the policy from the logits assigns non-zero probability to every action. Although individually rare, an opponent who prolongs the game by avoiding to get his piece captured and making a long series of neutral waiting moves will eventually benefit from one of these low-probability errors. In order to alleviate this issue we fine-tune the training with a different projection that thresholds and discretizes the action probabilities. The policy is written $\pi_{\theta, \epsilon_\text{tres}, n_\text{disc}}$ where $\epsilon_\text{tres}$ is the threshold level and $n_\text{disc}$ is the number of probability quanta used. This new projection of the policy is used to define new value function estimate and $Q$-functions estimate as $\hat v^1_{t,n}, \hat v^2_{t,n}, \hat Q^1_{t,n}, \hat Q^2_{t,n} = \Upsilon(\mathrm{T}_n,  \pi_{\theta_{n, \textrm{target}} \epsilon_\text{tres}, n_\text{disc}}, v_{\theta_{n, \textrm{target}}})$ that will be used instead of the previous estimate without any change to the rest of the 3 steps of the algorithm.

The parameters used to train \agent are summarized in Table~\ref{table:parameters_algo}.
\subsubsection*{Infrastructure for learning}

The IMPALA architecture ~\cite{espeholt2018impala} is adapted to the needs of the \reled algorithm, namely storing full episodes in the replay buffer. The learner reads a batch of full episodes, splits them into the sequence of ordered chunks of fixed length (mini batches), and computes the parameter updates on that sequence in a reverse order. This allows computing the exact full returns and to carry over the necessary information between mini-batches (see \reled algorithm above).

\begin{table}
\begin{center}
\begin{tabular}{ c|c } 
 \toprule
 parameter & value \\
 \midrule
Reward transform:  \\
 $\eta$  & $0.2$ \\
 $\Delta_m$  & 10k for $m\leq 100$,\\
 & 100k for $100<m\leq 165$,\\
 & 35k for $m>165$ \\
 max number of steps & 7.21M steps\\
 $lr_n$  & 0.00005 \\
 $c_{\textrm{clip gradient}}$  & 10000 \\
NeuRD parameters:  \\
 $\beta$  & 2.0 \\
 $c_{\textrm{clip NeuRD}}$  & 10000 \\
adam parameters:  \\
 $b_{1, \textrm{adam}}$  & 0.0\\
 $b_{2, \textrm{adam}}$  & 0.999 \\
 $\epsilon_{\textrm{adam}}$  & $10^{-8}$ \\
target network parameters:  \\
 $\gamma_{\textrm{averaging}}$  & 0.001 \\
$v$-trace parameters:  \\
 $\bar\rho$  & 1.0 \\
 $\bar c$  & 1.0 \\
Trajectory parameters:  \\
 $t_{\textrm{max}}$  & 3600 \\
 Batch size (number of trajectories per step) & 768\\
Fine tuning parameters \\
$\epsilon_\text{tres}$ & 0.03\\
$n_\text{disc}$ & 32\\
 \bottomrule
\end{tabular}
\end{center}
\caption{Parameters used during the training of \agent.}
\label{table:parameters_algo}
\end{table}

\subsection*{Game Rules and Neural Network Input Representation}

\subsubsection*{Basic Rules}
\label{sec:game_rules}

\begin{figure}[ht]
    \centering
    \begin{subfigure}[t]{0.32\textwidth}
        \includegraphics[width=\textwidth]{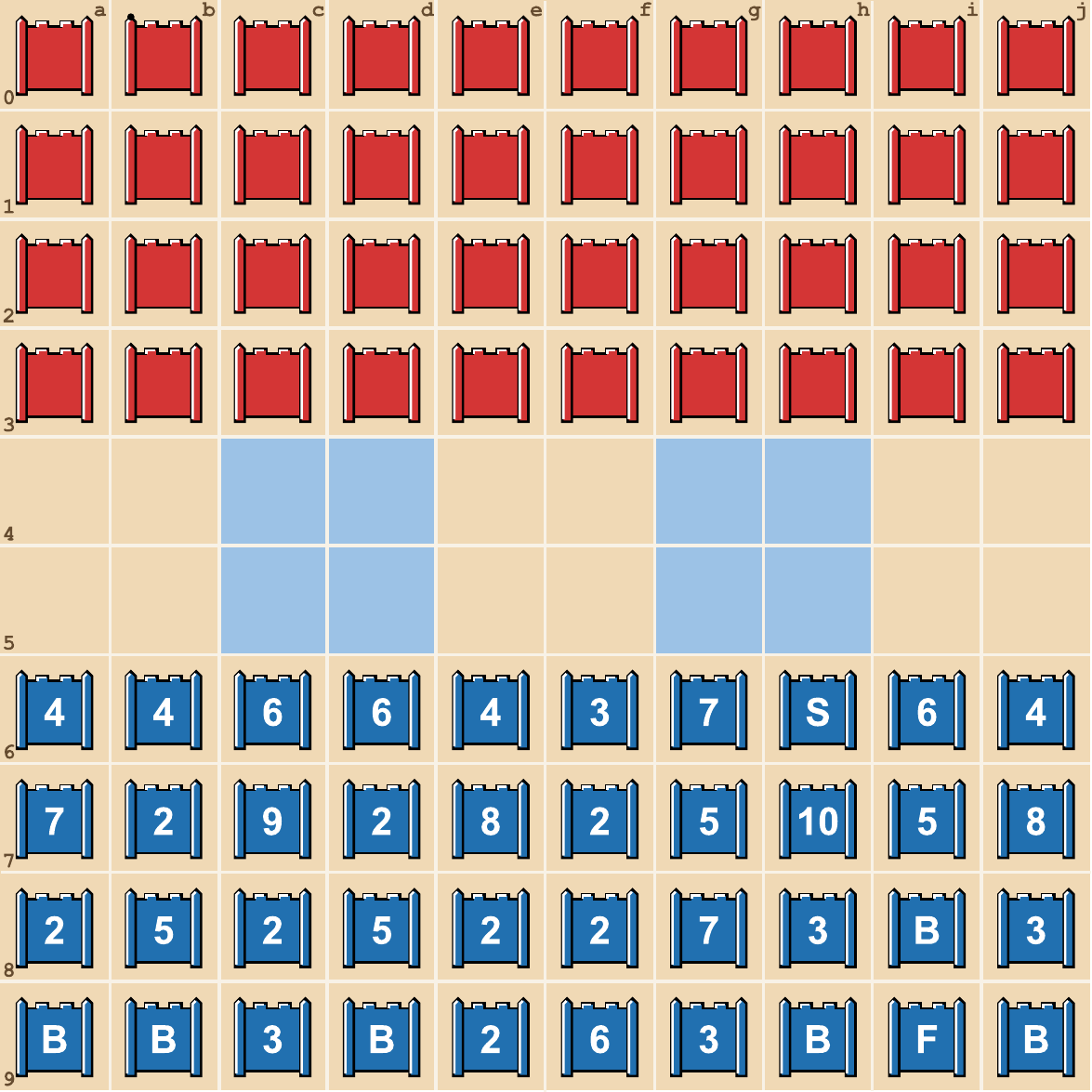}
        \caption{}
        \label{fig:example_deployment}
    \end{subfigure}
    \hfill
    \begin{subfigure}[t]{0.32\textwidth}
        \includegraphics[width=\textwidth]{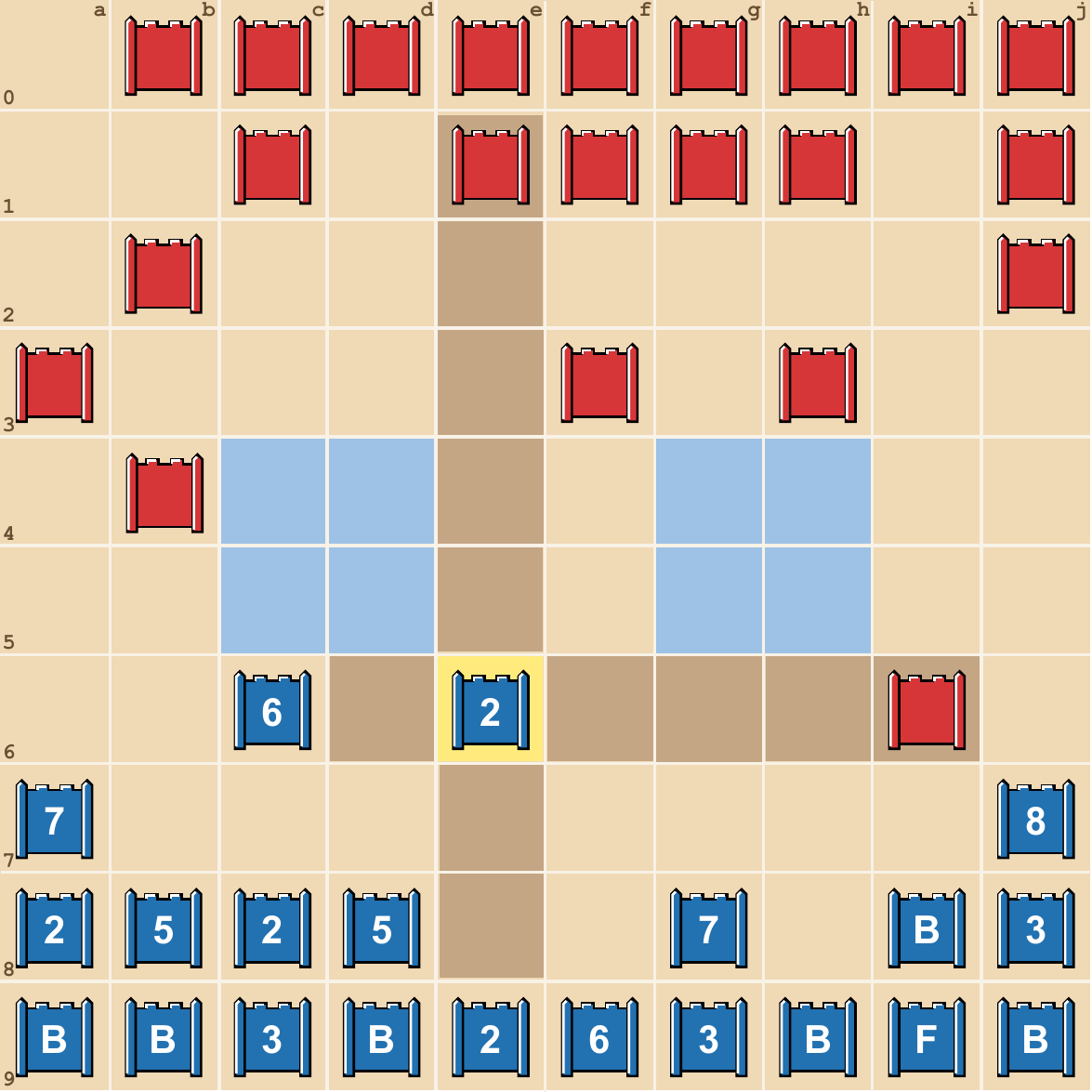}
        \caption{}
        \label{fig:example_move}
    \end{subfigure}
    \hfill
    \begin{subfigure}[t]{0.32\textwidth}
        \includegraphics[width=\textwidth]{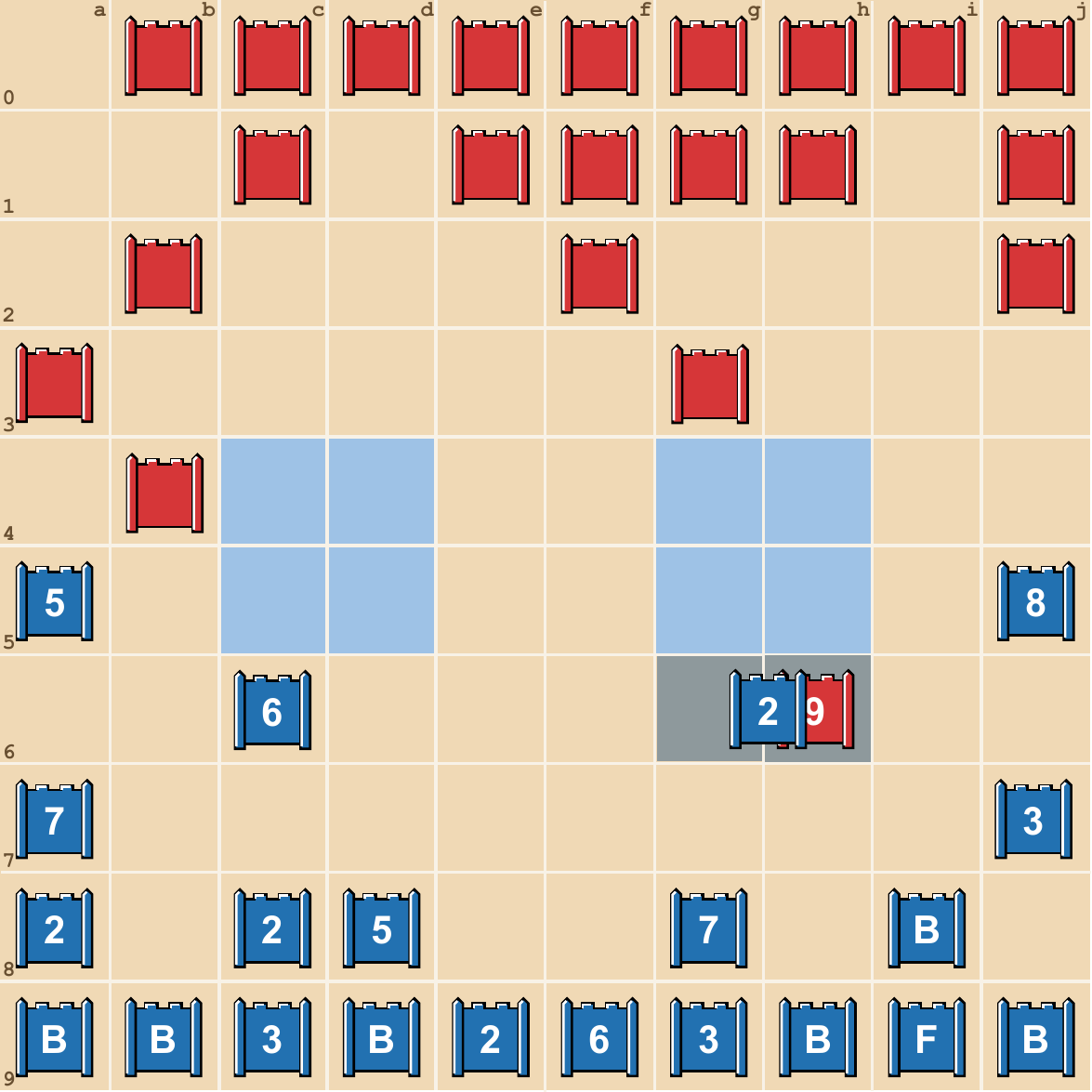}
        \caption{}
        \label{fig:example_attack} 
    \end{subfigure}
    \caption{The phases of a Stratego game: (a) the deployment phase during which players put their pieces on the board in a private configuration. (b) the game phase, where players alternate turns in moving pieces. Here a player chose to attack an opponent piece with the Scout (2), taking advantage of the long-range Scout action (c). As a result, both pieces are revealed; the Scout (2) is defeated by the General (9) and removed from the board.}
    \label{fig:board_examples}
\end{figure}

Stratego is a two-player board game, played between red and blue, corresponding to the colors of their pieces. The game board is a $10 \times 10$ grid of squares, with two $2 \times 2$ `lakes' which pieces may not move on or through - shown in light blue in Figure~\ref{fig:board_examples}. Each player starts with 40 pieces of 12 different types shown in the table below:

\begin{center}
\begin{tabular}{ cclc } 
 type & symbol & name & count \\
 \hline
0 & F &  Flag & 1 \\ 
1 & S & Spy & 1 \\ 
2 & 2 & Scout & 8 \\ 
3 & 3 & Miner & 5 \\ 
4 & 4 & Sergeant & 4 \\
5 & 5 & Lieutenant & 4 \\
6 & 6 & Captain & 4 \\
7 & 7 & Major & 3 \\
8 & 8 & Colonel & 2 \\
9 & 9 & General & 1 \\
10 & 10 & Marshal & 1 \\
11 & B & Bomb & 6 \\
 \hline
\end{tabular}
\end{center}

The game consists of two phases: the deployment phase, and a play phase. During the deployment phase --- as shown in Figure~\ref{fig:example_deployment} -- both players independently and privately position their 40 pieces on their side of the board in a $4 \times 10$ rectangular area, in any
configuration they choose.
In the play phase, the players take turns to move one of their pieces, starting with the red player.
The goal of the game is to either capture the opponent's Flag or capture all of their movable pieces.

On each player's turn, the current player moves one of their pieces either to an empty square or to a square occupied by one of the opponent's pieces, attacking it. It is not possible to move onto a lake or onto a square already occupied by one of the player's own pieces. Pieces can only move up, right, down, or left and cannot move diagonally. All pieces besides the Scout, Flag, and Bomb can move only one square at a time to one of their adjacent squares. The Scout can move through any number of empty spaces in one of the four directions. The Flag and Bombs remain static and as such cannot be moved.

When an attack is initiated, both players reveal their piece's type. The higher-valued piece remains on the board while the lower-valued piece is captured and removed from the board. If the pieces are of equal value, both are removed. There are two exceptions to this rule: Miners are able to capture Bombs, and Spies are able to capture Marshals if the Spy attacks the Marshal. A more detailed description of game rules, including provisions against repetition and endless chases, can be found in the official rule book \cite{stratego_rules}.

\subsubsection*{Drawing rules}
Under normal game rules, players can declare a game a draw by mutual agreement. We do not offer such negotiation actions to our agents, but instead trained and evaluated with the following (non-standard) rules: a game automatically ends in a draw either after 2000 total moves, or 200 consecutive moves have been made without any piece being attacked.

\subsubsection*{Agent action space}
The action space for the agent is discrete and of size 100 throughout the game, each action corresponding to a square on the board. The interpretation of an action and the definition of which actions are legal in a certain state depends on the phase of the game.

During the deployment phase, the player needs to \lq play' exactly 40 actions, one for each piece that is being deployed on the board as shown in Figure~\ref{fig:example_deployment}. The piece deployment order is fixed and sorted by piece-type: Flag (F), Bomb (B), Marshal (10), General (9), Colonel (8), Major (7), Captain (6), Lieutenant (5), Sergeant (4), Miner (3), Scout (2) and Spy (S). Thus each player first deploys their Flag, then their six Bombs, etc.

Deployment ends when both players have deployed all their pieces, entirely filling their four deployment rows. The play phase then commences with red making the opening move.

During the play phase, a full move is decomposed into two actions.
First, the agent selects the current location of one of its own pieces. Second, the agent selects a legal destination location for the selected piece as shown in Figure~\ref{fig:example_move}. If the new location is empty, the selected piece is moved there. If the new location contains an opponent piece, it initiates an attack, which is resolved as described in the section on the game rules, and illustrated in Figure~\ref{fig:example_attack}. Note that this two-step action decomposition affects both the observation and the agent design as detailed in the following subsections.

\subsubsection*{Neural Network Input Representation}
\label{sec:observation}

\newcommand{\obsprivateinfo}{\textbf{Prv}}
\newcommand{\obspublicinfo}{\textbf{Pub}}
\newcommand{\obsmoves}{\textbf{Mov}}

When played as a physical board game, human players observe the game as a sequence of
board positions where they see both the location and type of their own pieces, but only
the location of the opponent's pieces. The type of a piece is disclosed when it attacks
or is attacked, but this needs to be memorized by the player. Similarly, when a piece
moves, the information that it cannot be a Bomb or a Flag must be deduced by the human players
and memorized.

In common with many online versions of the game, we chose to present \agent with richer
observations which include the most important pieces of information available from the
prior play. For example, if an opponent piece has moved by more than square, the network
inputs will reflect that it could only be a Scout. Note that these deductions are policy
independent, and exclude heuristics which  depend on the opponent’s policy and other
higher level strategic choices.

Let a \textbf{piecetype assignment} for player $i$ be an assignment of a specific
piecetype to every piece of that player still on the board. It can be represented by a
$10 \times 10 \times 12$ tensor, where the two leading dimensions correspond
to a row ($r$) and column ($c$) of the Stratego board and such that $T_i[r, c, t] = 1$
if the square $(r, c)$ holds a piece of color $i$ of type $t$, else the value is 0.
For instance, $T_1[5, 6, 1]$ indicates that player red ($i=1$) has a Spy ($t=1$) at
location $(5;6)$

The \textbf{private information tensor} of player $i$, $\obsprivateinfo_i$ is the
piecetype assignment tensor of that player corresponding to the actual state of the
board for that player.

The \textbf{public information tensor} of player $i$, $\obspublicinfo_{i}$, is a
$10 \times 10 \times 12$ tensor such that $\obspublicinfo_{i}[r, c, t]$ is the probability
that square $(r,c)$ holds a piece of player $i$ of type $t$, under uniform sampling of
piecetype assignments consistent with the move sequence of the game so far. The requirement
here is just that the move sequence would be legal, not that it would be rational or
in accordance with some policy.

This public information tensor has the property that $\sum_t \obspublicinfo_{i}[r, c, t]$ = 1
for each square $(r, c)$ occupied by $i$, and that $\sum_{r, c} \obspublicinfo_{i}[r, c, t]$
is the number of pieces of player $i$ of type $t$ still on the board. Given there are only
two categories of unknown pieces (moved and non-moved), this public information tensor can
be computed efficiently as follows:
\begin{equation}
    \text{\obspublicinfo}_{i}[r, c, t] =
    \begin{cases}
      0 & \text{if there is no piece belonging to player $i$ at position $(r, c)$} \\
      1 & \text{if the piece at $(r, c$ is known to have type $t$} \\
      \frac{\#\text{unrevealed}(t)}{\sum_{k \ne \text{B}, \text{F}} \#\text{unrevealed}(k)} & \text{if the piece at $(r, c)$ has ever moved and $t \ne \text{B}, \text{F}$} \\
      \frac{\#\text{unrevealed}(t)}{\sum_{k} \#\text{unrevealed}(k)} & \text{if the piece at $(r, c)$ has never moved}
    \end{cases}
\end{equation}
Where $unrevealed(t)$ is the number of pieces of player $i$ of type $t$ which are still on the board and where
the type of the piece hasn't been revealed by an attack or by making a Scout move.

A move $m$ during a game observed by player $i$ is encoded as a $10 \times 10$ tensor $\obsmoves_i^m$ such that:
\begin{equation}
	 \text{\obsmoves}_{i}^{m}[r, c] =
	 \begin{cases}
	 	 -1 &\text{if the piece made a regular move from square $(r,c)$} \\
	 	 -(2 + t/12) &\text{if a piece of type $t$ attacked from square $(r,c)$} \\
	 	 1 &\text{if the piece moved to or attacked square $(r,c)$}	 \\
	 	 0 &\text{elsewhere}
	 \end{cases} 
\end{equation}

In the remainder, when referring to a player $i \in \{\text{red}, \text{blue}\}$, we use $-i$ to denote the opponent. The observation for player $i$ consists of the components shown in table \ref{table:observation}. These are stacked into a $10 \times 10 \times 82$ tensor by expanding the scalar components to a $10 \times 10$ tensor with constant value.

\begin{table}
\begin{center}
\begin{tabular}{ | m{12cm} | m{3cm} | } 
\hline
observation component & shape \\
\hline\hline

The lakes on the board, where a square that is a lake has value 1, otherwise 0. & $10 \times 10 $ \\
\hline

The player's own private information $\obsprivateinfo_i$ & $10 \times 10 \times 12$ \\
\hline

The opponent's public information $\obspublicinfo_{-i}$. Contains all 0's during the deployment phase. & $10 \times 10 \times 12$ \\
\hline

The player's own public information $\obspublicinfo_{i}$: this informs $i$ on the information $-i$  has on $i$'s pieces. Contains all 0's during the deployment phase. &  $10 \times 10 \times 12$ \\
\hline

An encoding of the last 40 moves: $\obsmoves_i^m$ for each move made up to 40 steps ago. & $10 \times 10 \times 40$ \\
\hline

The ratio of the game length to the maximum length before the game is considered a draw. & scalar \\
\hline

The ratio of the number of moves since the last attack to the maximum number of moves without attack before the game is considered a draw. & scalar \\
\hline

The phase of the game: either deployment (1) or play (0). & scalar \\
\hline

An indication of whether the agent needs to select a piece (0) or target square (1) for an already selected piece. 0 during deployment phase. & scalar \\
\hline

The piece selected in the previous step (1 for the selected piece, 0 elsewhere), if applicable, otherwise all 0's. & $10 \times 10$ \\
\hline

\end{tabular}
\end{center}
\caption{The components of the agent observation. These are stacked into a $10 \times 10 \times 82$ tensor by expanding the scalar components to a $10 \times 10$ tensor with constant value.}
\label{table:observation}
\end{table}

\subsubsection*{Player-centric observations and actions}
To facilitate training a single agent that can play both sides of the board, the observation is presented player-centric: the tensor for player blue is rotated 180 degrees. Likewise, the interpretation of the actions (between 0 and 99) as squares on the board are also done relative to the side of the board of the player.

\subsection*{Network Architecture}

\begin{figure}[ht!]
    \centering
    \includegraphics[width=1\textwidth]{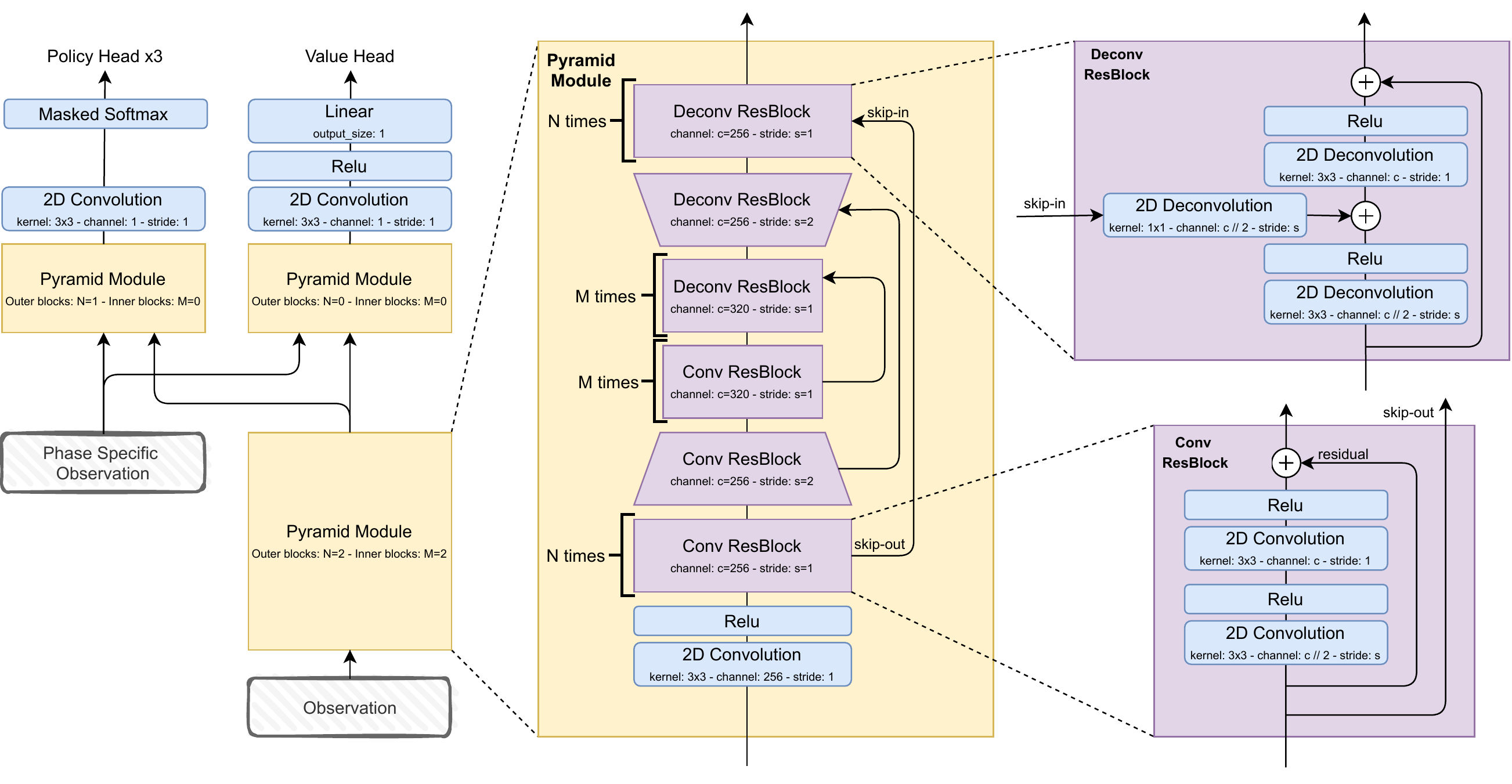}
    \caption{Network implementation details. When applying striding, residual connections are also processed by a convolution layer with $1\times1$ kernel (hidden for clarity).}
    \label{fig:network}
\end{figure}

\paragraph{Network Modules}
The agent policy is parameterized through a deep neural network. A torso module first processes the observation to produce a board game embedding, represented as a tensor of spatial dimension $10\times10$ and channel dimension $256$. This representation is then provided to three policy head modules specialized by game phase and a value-head module. 

First, the deployment head is used during the deployment phase, i.e. the 40 first steps of the game. It takes as input the board game embedding. It then outputs a probability distribution of dimension $10\times10$ over the board (restricted to legal positions) to indicate the emplacement of the next piece to deploy. The piece deployment always follows the same order, so we did not provide complimentary observation to the board game embedding.

Second, the piece-selection head is used during the first stage of the game phase, i.e., selecting the unit to play. It takes as input the concatenation of the board game embedding and the tiled no-attack ratio. 
It then outputs the probability distribution of dimension $10\times10$ (restricted to playable units) to pick the unit to play. This distribution is then sampled to proceed to the second stage of the game phase.

Third, the piece-displacement head is used during the second stage of the game phase, i.e., moving the selected piece and potentially attacking an opponent unit. It takes as input the concatenation of the board game embedding, the no-attack ratio, and the one-hot representation of the selected piece. This one hot representation is a sparse tensor of spatial dimension $10\times10$ and channel dimension $10$ that respectively encode the spatial location of the selected piece and the unit-type across movable units (all units but the Bomb (B) and the Flag (F)).

Fourth, the value head is used during the training to compute the value function of the agent. It takes as input the concatenation of the board game embedding, the no-attack ratio, and the one-hot representation of the selected piece. Note that the no-attack ratio is zeroed-out during deployment, and the one-hot representation of the selected piece is zeroed-out during deployment and the first stage of the game phase. While we could use a value head for each policy head, we used a single value head for the three policies to reduce the agent's memory footprint. This zeroing is then necessary to have a fixed input shape across all phases.

\paragraph{Neural Implementation}
All neural modules, i.e. the torso and heads, are based on a pyramid-like architecture~\cite{lin2017feature}, which are themselves composed of diverse resblock modules~\cite{he2016deep}. Note we refer as inner blocks, the blocks that perform spatial dimension reduction through convolution layers~\cite{lecun1998gradient}, and outer blocks that spatially upscale the intermediate feature map through deconvolution layers~\cite{long2015fully}.

\noindent The pyramid module is constructed as follow from input to output: 
\setlist{nolistsep}
\begin{itemize}[noitemsep]
    \item $1$ convolution layer with $C=256$ channel, no striding, $3\times3$ kernel and relu.
    \item $N$ outer-convolution resblocks with $C=256$ channels and no striding.
    \item $1$ strided-convolution resblock with $C=256$ channels and $S=2$ striding. 
    \item $M$ inner-convolution resblocks with a channel size $C=320$ and no striding.
    \item $M$ inner-deconvolution resblocks with a channel size $C=320$  and no striding.
    \item $1$ strided-deconvolution resblock with $C=256$ channels and $S=2$ striding. 
    \item $N$ outer-deconvolution resblocks with $C=256$ channels and no striding.
\end{itemize}
Pyramid modules also include skip-connections that go from the convolution resblock to its symmetric deconvolution resblock. Policy heads use $N=1$ outer resblocks, strided-convolution resblocks and no inner resblocks $M=0$. The final deconvolution is followed by a 2D-convolution layer with a single channel, no striding, $3\times3$ kernel, and relu activation to output the action logits. Forbidden actions are then masked by turning their logits to $-\infty$, and the remaining logits are turned into a probability distribution through a softmax activation layer. The value head uses no outer resblocks $N=0$, the strided-convolution resblocks and no inner resblocks $M=0$. The final deconvolution is followed by a 2D-convolution layer with a single channel, no striding and $3\times3$ kernel, and relu activation. The logits are then flattened and processed by a linear layer to output the final value-function scalar.

\noindent The convolution resblocks are constructed as follow from input to output: 
\setlist{nolistsep}
\begin{itemize}[noitemsep]
    \item A skip-out connection that is fed back to the symmetric deconvolution resblock.
    \item A convolution layer with $C // 2$ channel, optional $S$ striding, $3\times3$ kernel and relu.
    \item A convolution layer with $C$ channel, no striding, $3\times3$ kernel and relu.
    \item A residual connection that sum the initial resblock input and the current logits.
\end{itemize}

\noindent The deconvolution resblocks are constructed as follow from input to output: 
\setlist{nolistsep}
\begin{itemize}[noitemsep]
    \item A deconvolution layer with $C // 2$ channel, optional $S$ striding, $3\times3$ kernel and relu.
    \item A skip-in connection that sum the symmetric skip-out connection and the current logits. The skip-connection is first processed by a 2D-deconvolution layer with $C//2$ channel, optional $S$ striding, $1\times1$ kernel and no activation.
    \item A deconvolution layer with $C$ channel, no striding, $3\times3$ kernel and relu.
    \item A residual connection that sum the initial resblock input and the current logits.
\end{itemize}
When performing striding in resblocks, the residual connection is completed by a 2D-convolution layer with $2$ stride, $1\times1$ kernel and no activation to fit the dimension change.

\subsection*{Infrastructure and Setup}

\begin{figure}[th!]
    \centering
    \includegraphics[width=1\textwidth]{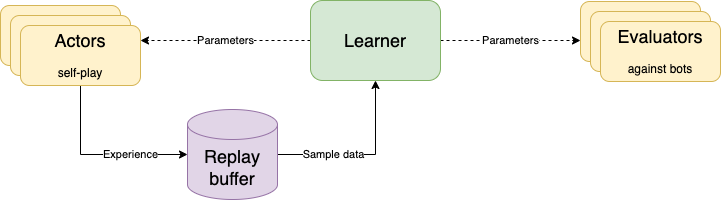}
    \caption{Agent decomposition. Learner samples games/trajectories from Replay Buffer, improves weights, and periodically distributes these to actors and evaluators. Actors self-play and write experience into Replay Buffer. Evaluators play against other bots to assess performance.}
    \label{fig:InfraActorsLearner}
\end{figure}

The \agent training pipeline follows Sebulba Podracer architecture from \cite{hessel2021podracer}. It consists of Actors, Replay Buffer, Learner and Evaluators (Figure~\ref{fig:InfraActorsLearner}).

\begin{itemize}
    \item[*] The Actors self-play and write full games into Replay Buffer.
    \item[*] The Replay Buffer, as a queue, stores games from Actors until the Learner reads it.
    \item[*] The Learner, extracts a batch of games from Replay Buffer, improves
    the network weights, and periodically sends those to Actors/Evaluators.
    \item[*] The Evaluators play against fixed opponent bots, e.g. Uniform or Demon of Ignorance; then plot the aggregated statistics.
\end{itemize}

\paragraph{Deployment}

We deploy Actors, Learner and Buffer together on the same machine, since they all use/need
different kinds of resources:

\begin{itemize}
    \item[-] Actors run environment and inference and use CPU and a small amount of accelerator FLOPs.
    \item[-] Replay Buffer predominantly uses RAM.
    \item[-] Learner heavily uses accelerator FLOPS.
\end{itemize}

Evaluators run separately as a flock of low-priority machines.

\paragraph{Scalability}

\begin{figure}[th!]
    \centering
    \includegraphics[width=1\textwidth]{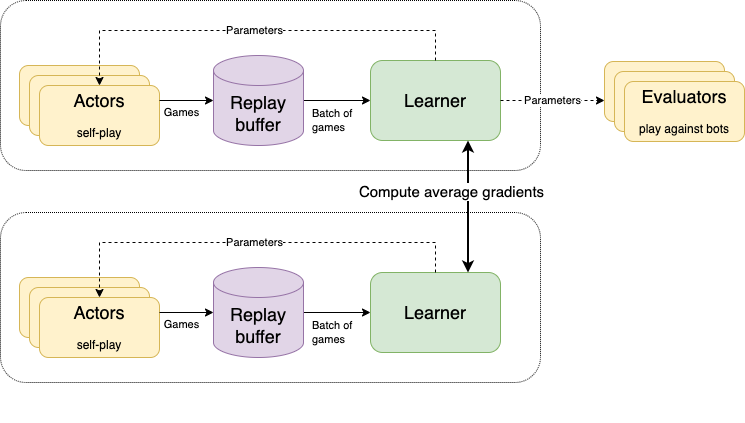}
    \caption{Several machines participate in training, each embedding its own Learner, Actors and Replay Buffer. At each parameter updates, the gradients are averaged across all machines. Everything else is as if machines were independent.}
    \label{fig:InfraSIMD}
\end{figure}

To scale the training pipeline we rely on SIMD model (Figure~\ref{fig:InfraSIMD}) implemented in JAX~\cite{jax2018github} as `pmap` transformation.

Briefly on SIMD, several learner machines participate in training; each does a learning step in
sync with other machines; each reads a batch of different games, computing network
parameter gradients; computes the average of the gradients across all machines; and finally
each applies those average gradients, as updates, to network weights.

To generate the batches of games, each learner machine runs, as separate threads, its own
Actors and a single Replay Buffer.

To train the final agent we used 768 TPU nodes used for Learners and 256 TPU nodes for Actors.

\paragraph{Differences from Sebulba Podracer}

\begin{itemize}
    \item The Actors environment-agent loop is implemented in C++ using fibers and
    the native open spiel~\cite{LanctotEtAl2019OpenSpiel} interfaces. This provides better overall throughput for
    playing games.
    \item Full games are stored in the Replay Buffer, and each Learner step consumes
    a batch of full games. Note that the game length is variable. See below for the details.
    \item The evaluators are run a separate flock of low-priority machines, that evaluate
    the agent against opponent bots. It decouples opponent bot' hardware requirements from
    the main training pipeline.
\end{itemize}

\paragraph{Full games learning}

To compute the exact returns on a variable-length trajectory, we store full games in the Replay Buffer (instead of a fixed length sequences of steps as in most RL agents). When learner samples a batch of games, shorter games are padded to the length of the longest game in the batch. The batch is chopped into a variable number of fixed length chunks along the time-axis. The learner then does two passes over this list of chunks: a) a forward pass to recompute the exact network (LSTM) state for each timestep and trajectory statistics; b) a backward pass, where each chunk is reprocessed again, but is given the exact network state at the beginning of the chunk and full trajectory statistics, such as the exact returns, etc.

Note that (a) happens outside of gradients computation, and serves to avoid bootstrapping or any kind of approximation for both the returns and for the network state. While (b) happens within the loss and serves to compute gradients for network weights.

The gradients computed on each chunk during (b) are accumulated, and applied using optimizer onto the network weights at the end of the (b) pass.

\begin{figure}[th!]
    \centering
    \includegraphics[width=1\textwidth]{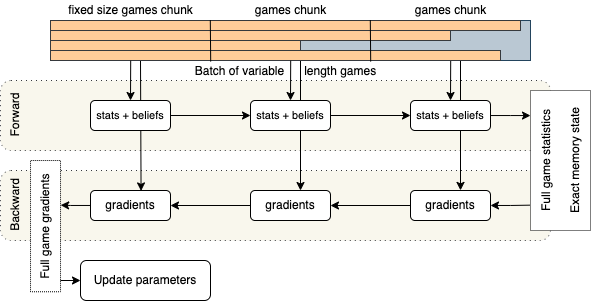}
    \caption{Variable length games are padded and split into fixed size chunks. Chunks
    are forward processed to compute global statistics and up-to-date beliefs/memory.
    Then chunks are processed in reverse order to compute parameter gradients. Finally
    accumulated gradients are used to update network weights.}
    \label{fig:Multipass}
\end{figure}

\subsection*{Test Time Improvements}
\label{sec:testtimeimprovements}
When evaluating our Stratego agent, we make some adjustments to the policy given by the trained neural network to improve the strength of the agent further and remove
some non-performance-improving characteristics of its play which are annoying to human opponents.

\subsubsection*{Post-processing of policy}
Given Stratego is a complex imperfect information game, we expect the Nash equilibrium policy to be non-deterministic in many states of the game. We indeed observe that the policy produced by the \reled algorithm is non-deterministic in most states. However, the policy often gives a small but non-zero probability to moves which are clear blunders, and which would result in the agent losing a game which it should have won.
Since Stratego games can take up to 2,000 moves, playing according to this stochastic policy will result in taking several very-low-probability actions over the course of the game.
This is especially true if the opponent is aware of the issue and prolongs the game by avoiding attacks and making neutral waiting moves.
Presumably an exact equilibrium strategy would give a probability of precisely zero to these actions -- we therefore apply a heuristic post-processing of the policy to eliminate very-low-probability moves, without making the policy of the agent too deterministic, which would make it exploitable:
\begin{itemize}
    \item Thresholding: all actions with probability lower than a fixed threshold $\epsilon_\text{tres}$ are dropped and the policy is renormalized. If no actions with positive probability would remain, the policy is left unchanged.
    \item Discretization: sorted from high to low, probabilities are rounded up to the nearest multiple of $1/n_\text{disc}$ and remaining weights are discarded once a sum of 1 is reached.
\end{itemize}
Table \ref{table:ttisettings} lists the values of these parameters we used.

\newcommand{\eagerness}{\alpha_\text{eag}}

\subsubsection*{Avoid repetitive play}
We observed that our agent often repetitively moved a piece back-and-forth on two squares, thereby threatening a known lower-valued opponent piece. While this is allowed by the rules up to a certain limit (see the two-square and more-square rules in \cite{stratego_rules}), the repetition is annoying for human opponents.\footnote{In some cases it could even be considered as "unsporting behavior" according to the ISF rules\cite{stratego_rules}. } Moreover we observed in some cases in human evaluation that this style of play resulted in a draw where the agent actually could have won. We apply two techniques to our agent that avoid this style of play:
\begin{itemize}
\item Avoid pointless threats: initially the agent is allowed to threaten an opponent piece by moving move back and forth three times between two adjacent squares, in accordance with the two-square rule. After that, for the same two pieces on the same 2x2 region of the board, only a single threatening move is allowed: if the opposing piece retreats as before, the agent cannot immediately move to re-threaten it unless the policy has no alternative moves.
\item Eagerness: decreases the agent's perception of how much time remains before the game will be considered a draw due to no no piece having been attacked. As explained in the section on Neural Network Input Representation, the observation presented to the agent contains a ratio $r$ indicating how close the game is to being declared a draw, which is reset to zero every time an attack takes place, and reaches the value $1$ when a draw is declared. In order to induce the agent to be more eager in its play, at test time we modify this to $r' = 1 - (1 - r)^{\eagerness}$ with $\eagerness$ a parameter defined in table \ref{table:ttisettings}.
\end{itemize}
Note that the pointless threat restriction applies only to our agent, not to its opponent.

\subsubsection*{Memory heuristic}
In our human evaluation it was observed that in some cases the observation presented to the neural network, as described in the section on the Neural Network Input Representation, does not contain sufficient history to allow a detailed assessment of a Stratego state. This is an area we think has considerable potential for further performance improvement,
by using agent architectures that can better cope with long memory. 

That being said, many of these memory problems were very similar and could be solved by tracking of a small number of move patterns. Specifically, our \emph{memory heuristic}
assumes that:
\begin{itemize}
\item A Spy would attack a known Marshal;
\item A Marshal would attack a known General or Colonel if no recapture is possible;
\item A General would attack a known Colonel if no recapture is possible;
\end{itemize}
And therefore that if a player instead makes a non-attacking move, then any piece that had the chance to attack according to the rules described is not of the corresponding type.
These eliminated possibilities are tracked per piece throughout the game and influence the policy by modifying the network input: the corresponding entries in both  $\obspublicinfo_{i}$
and $\obspublicinfo_{-i}$ are set to 0. 

The assumptions of this heuristic are very often, but not always, met, even in self-play. Despite the occasional inaccuracies, overall using this heuristic seems to improves the performance in matches against humans. A qualitative observation is that it mostly acts through $\obspublicinfo_{i}$, i.e. by more accurately tracking the information the
human has on the agent's pieces, for example by avoiding some types of inconsistent bluffing. 

\subsubsection*{Value bounds heuristic}
While the policy post-processing described above removes most errors without making use of game-specific logic, the evaluation of successive agent checkpoints with Vincent de Boer revealed that occasionally an obvious mistake still had support in the policy and if played, could decide a game. We therefore added a \emph{value bounds heuristic} which aims to eliminate obvious mistakes by leveraging the value function which is trained alongside the policy, and which can be
used to score any state where the agent needs to act. Two-step lookahead is performed to estimate an \emph{upper bound} of the value of each action in the policy. These values are not informative about the relative merits of actions considered, but if an estimated upper bound for an action is \emph{lower} than the value of the current state, then the action is considered to be an error, and removed from the policy.

In performing this evaluation, no probabilistic assessment is made of the opponent's private information; instead the best-case private information of the
opponent is assumed for the purpose of evaluating attacks, and the worst-case response by the opponent from a set of safe actions.
\newcommand{\neuralvalue}{v^\textrm{nn}}
More precisely (using terminology and notations introduced in the section on Imperfect Information Games) let $\neuralvalue(x)$ be the value function provided by the neural network applied to information state $x$.
We define an action (consisting of a piece moving or attacking) to be \emph{safe} if all of the following conditions are met: 
\begin{itemize}
\item the piece is already known to be movable,
\item in case of a long move/attack, the piece already known to be a Scout,
\item in case of an attack, it is a guaranteed win, for any private info of both players consistent with the public info,
\item the piece cannot be subsequently captured, for any private info of both players consistent with the public info
\end{itemize}
\newcommand{\safeactions}{\mathcal{A}_\textrm{safe}}
\newcommand{\valueupperbound}{\widehat{v^\textrm{nn}}}
So safety is defined based only on public information of both players. We write $\safeactions(h)$ for the  set of all safe actions at history $h$. Given an information state $x$, we define the value upper bound for action $a$ as 
\begin{equation}
\valueupperbound(x, a) = \max_{h \in x} \min_{a' \in \safeactions(ha)} \neuralvalue(x(haa'))
\label{eq:valueupperbound}
\end{equation}
If $\valueupperbound(x, a) + \epsilon_\text{vb} < \neuralvalue(x)$ then $a$ is removed from the policy, if it supports alternative actions, with $\epsilon_\text{vb}$ a margin defined in table \ref{table:ttisettings}.
Equation \ref{eq:valueupperbound} does not deal with terminal states for simplicity but can be generalized to these cases straightforwardly. In practice $\valueupperbound(x, a)$ can be determined efficiently without enumerating all $h \in x$ because at most two unknown opponent pieces can be involved: a piece attacked by the agent and subsequently a piece attacking the agent.

This heuristic only occasionally intervenes; for example in the matches played on Gravon, it affected less than 1.5\% of \agent's turns. Also, it is interesting to note that both this heuristic and the memory heuristic do not significantly improve the winrate when \agent is evaluated against a version of itself without these heuristics. They only empirically seem to avoid some mistakes observed in matches against humans. 

\begin{table}
\begin{center}
\begin{tabular}{ ccl } 
\hline
$\epsilon_\text{tres}^\text{(deploy)}$ & 0.03 & threshold used in deployment phase\\
$\epsilon_\text{tres}^\text{(play)}$ & 0.03 & threshold used in play phase \\
$n_\text{disc}^\text{(deploy)}$ & 32 & discretization used in deployment phase \\
$n_\text{disc}^\text{(play)}$ & 16 & discretization used in play phase \\
$\eagerness$ & 2.0 & eagerness parameter \\
$\epsilon_\text{vb}$ & 0.05 & margin used by the value bounds heuristic \\
\hline  
\end{tabular}
\end{center}
\caption{Test time improvement settings.}
\label{table:ttisettings}
\end{table}

\section*{Additional results}
\subsection*{Details on human evaluation on Gravon}
\agent was evaluated on Gravon, beginning of April 2022, with the consent of Thorsten Jungblut, the Gravon platform owner. This resulted in roughly 50 ranked matches. The ratings of \agent in the 2022 and all-time ranking were computed on April 22\textsuperscript{nd} 2022 and three matches were ignored for this computation: two were played with an earlier training snapshot, one timed out due to a human error.

\subsection*{Details on external Stratego program evaluation}

\textbf{Probe} is a program which was three-fold winner of the Computer Stratego World Championship (2007, 2008, 2010) \cite{stratego_wikipedia} and is currently available as the "Heroic Battle" app on Android. We evaluated version 2.0.37  using the Expert skill level and played the same number of games for the different playing styles available: Conservative, Moderate and Aggressive.

\textbf{Master of the Flag} is a program which won the Computer Stratego World Championship in 2009  \cite{stratego_wikipedia}. It is currently available on a website \cite{master_of_the_flag} and we tested against algorithm version 5.2.0.40. In the interface, Master of the Flag only plays as Blue so we evaluated in this setting only. 

\textbf{Demon of Ignorance} is an open-source implementation of Stratego with an accompanying AI bot, written in Java \cite{demon_of_ignorance}. We evaluated version 0.13.4. The bot's AI-level can be configured, which controls its thinking time. We observed that the performance of this bot against our agent saturates at an AI-level of 8, so this is the setting we use for evaluation. Occasionally this bot does not produce a valid action, if this happens 3 times in a row, the ongoing game is discarded.

\textbf{Asmodeus}, \textbf{Celsius}, \textbf{Celsius1.1}, \textbf{PeternLewis}, \textbf{Vixen} are agents that were submitted in 
an Australian university programming competition in 2012 \cite{ucc_australia2012}, won by \emph{PeternLewis}. These agents sometimes pick moves that violate the two- and more-square rule of Stratego. We therefore evaluated using a game variant that allows such repetitive moves, but we discarded matches that resulted in a draw due to such endless repetition. We evaluated at commit 54f0978 of the repository.

\section*{Quotes from Stratego Experts}

Thorsten Jungblut, owner of the Gravon platform:
\begin{quote}
Many players in the past thought that there will never be an AI for Stratego that could be a real competition for human players, or even play in the top ten. Obviously, they were wrong.
\end{quote}

Vincent de Boer, former Stratego world champion, evaluated \agent as follows:

\begin{quote}
    The level of play of DeepNash surprised me. I had never seen or heard of an artificial Stratego player that came close to the level needed to win a match against an experienced human player, but after playing against DeepNash myself I was not surprised by the top-3 ranking it later on achieved on the Gravon internet platform. I would expect this agent to also do very well if it participated in the World Championship.
\end{quote}
\clearpage

\bibliography{scibib}
\bibliographystyle{Science}

\end{document}